\documentclass{article} 
\usepackage[preprint]{tmlr}

\usepackage{amsmath,amsfonts,bm}

\def\eqref#1{equation~\ref{#1}}

\def\1{\bm{1}}

\DeclareMathAlphabet{\mathsfit}{\encodingdefault}{\sfdefault}{m}{sl}
\SetMathAlphabet{\mathsfit}{bold}{\encodingdefault}{\sfdefault}{bx}{n}

\usepackage{hyperref}
\usepackage{url}

\usepackage[utf8]{inputenc}
\usepackage[T1]{fontenc}
\usepackage{charter}
\usepackage{booktabs}
\usepackage{amsfonts}
\usepackage{nicefrac}
\usepackage{microtype}
\usepackage{xcolor}
\usepackage{graphicx}
\usepackage{amsmath, amssymb, amsthm, mathtools}
\usepackage{bm}
\usepackage{enumitem}
\usepackage{algorithm}
\usepackage{algorithmic}
\usepackage{caption}
\usepackage{subcaption}
\usepackage{multirow}
\usepackage{tcolorbox}
\usepackage{placeins}
\usepackage{subcaption} 
\usepackage{amssymb}
\usepackage{wrapfig}
\usepackage[table]{xcolor}

\hypersetup{colorlinks=true, linkcolor=blue!70!black, citecolor=green!50!black, urlcolor=blue!60!black}

\newcommand{\bF}{\bm{F}}
\newcommand{\bsig}{\bm{\sigma}}

\newcommand{\Id}{\bm{I}}

\title{FracGen: Learning How Objects Stretch and Tear with Physics-Informed Video Generation}

\author{\vspace{-20pt}
\\Trong-Tung Nguyen\quad\quad Jiahan Zhang \quad\quad Anand Bhattad \\
[5pt]
Johns Hopkins University \\
[10pt]
\normalfont \textbf{Project Website}: \href{https://fracgen.github.io}{\texttt{https://fracgen.github.io}}%
}

\begin{document}

\maketitle
\vspace{-0.8cm}
\begin{center}
    \includegraphics[width=0.95\linewidth, trim={0cm 2.8cm 0cm 3.3cm}, clip]{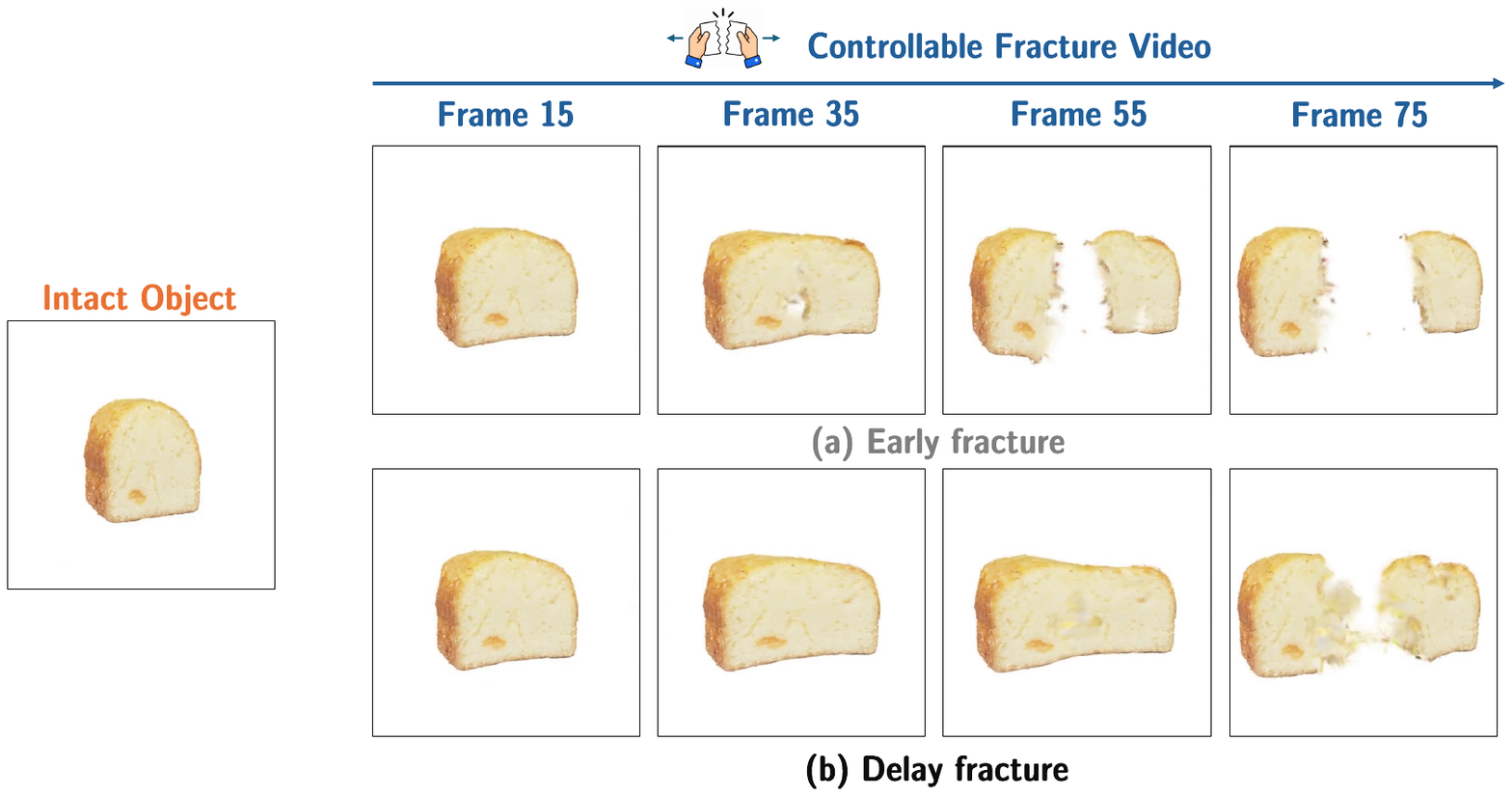}
    \captionof{figure}{\textbf{Controllable stretching and tearing with FracGen}. Given a single object image and specified force, material, and fracture conditions, FracGen generates a video of the object deforming and tearing. Under the same applied force, varying the fracture properties produces earlier tearing (top) or delayed tearing after greater deformation (bottom).} 
    \label{fig:teaser}
\end{center}

\begin{abstract}
We introduce \textbf{FracGen}, a fracture-aware video generation model that produces plausible, controllable fracture dynamics from a single image of an intact object, conditioned on physics signals. To train FracGen, we build \textbf{FracSim}, a fracture-aware simulation framework that augments material point method (MPM) simulation with a continuum damage model, producing paired fracture videos and dense, pixel-aligned physical fields at no additional cost beyond standard rendering. FracGen leverages these maps in two ways: it is trained to jointly predict them alongside RGB video, encouraging the model to capture physical state rather than surface appearance; and it is supervised with physics-informed losses that encourage consistency among the predicted maps. As a result, FracGen captures distinct material-specific fracture behavior without expensive test-time simulation or per-scene tuning, while offering fine-grained control over where an object tears, how fast the crack propagates, and how much deformation precedes failure. We further introduce a benchmark for evaluating the physical plausibility of generated fracture video, and show through extensive experiments that FracGen outperforms existing video generation baselines in both physical and visual fidelity. Results are best viewed in our project website: \url{https://fracgen.github.io/}.
\end{abstract}

\vspace{-0.5cm}
\section{Introduction}
\label{sec:intro}

Things break all the time. We drop, tear or snap objects and know what a fracture or damage looks like. Bread breaking like glass or a mug tearing apart like paper looks immediately wrong. Any convincing fracture-generation algorithm must match these physical intuitions — not just the final broken shape, but also how an object deforms, when it starts to fail and how the fracture develops.
 

A growing body of work shows that modern generative models internalize substantial knowledge of physical dynamics directly from data ~\citep{bhattad2023stylegan, du2023generative, liu2024physgen, physctrl2025, Xing2024luminet, gillman2026force}. In this paper, we ask whether this extends to fracture: can a video generator learn how objects break from limited data and lightweight fine-tuning? 

Recent work has begun to condition video generation on physical signals. ForcePrompting \citep{gillman2026force} shows that generators can respond to push/pull forces, but does not explicitly model damage accumulation or fracture. Physics-based approaches address different parts of this problem. PhysGaussian \citep{Xie_2024_CVPR} produces physically grounded deformation but lacks a damage model for fracture, while Fracture-GS \citep{wang2026fracturegs} introduces a Collision Material Point Method (Collision-MPM) to simulate fracture under extreme mechanical collisions. These simulation pipelines require a reconstructed scene and configured physical conditions. Our goal is to instead start from a single photograph and specified material and force conditions, and answer by generating \emph{``what happens if I pull here until it breaks?''}

To this end, we introduce \textbf{FracGen}, a video generation model that produces fracture dynamics for objects under tensile stretch. We focus on \emph{stretch-to-tear} fracture, where substantial deformation precedes failure, making the evolution before fracture an important part of the generation task. FracGen captures distinct material-specific behaviors, from objects that fail after little stretching to those that undergo large deformation before tearing apart.

Teaching a generator these behaviors requires supervision beyond visual appearance: RGB frames alone do not explicitly describe the mechanical state that leads to fracture. We observe that simulators like PhysGaussian already track rich physical information that is not exposed in their rendered videos. Each particle carries a deformation gradient, a $3\times3$ matrix encoding local deformation, from which strain can be computed and stress obtained through the material's constitutive law. Particle velocities further describe the local motion. We introduce a damage field that tracks the accumulated failure state of each particle and governs fracture. Rendering these quantities as pixel-aligned maps alongside RGB frames yields dense, physically grounded supervision without additional simulation runs. We call this augmented simulation framework \textbf{FracSim}, and its rendered physical fields \textbf{FracPhys Maps}.


We use these maps in two complementary ways, forming our main technical contributions. First, we extend the video generator to jointly predict FracPhys Maps alongside the fracture RGB video, forcing the model to encode the underlying mechanical state rather than surface appearance alone. These maps also enable us to inspect the physical state underlying the generated video. Second, we introduce physics-informed losses, derived from constitutive modeling, that encourage consistency among the predicted physical channels: the predicted strain, stress, and damage representations are coupled through a latent consistency loss inspired by constitutive modeling and stress degradation.
We further introduce a benchmark targeting the evaluation of physical alignment in generated fracture videos, along with extensive experiments comparing FracGen against baseline methods to quantify its effectiveness. In summary, our contributions include:

\vspace{-5pt}
\begin{enumerate}[leftmargin=1.8em, itemsep=0pt]
  \item \textbf{FracSim}, a fracture-aware simulation framework that incorporates continuum
  damage mechanics into MPM, producing paired fracture videos and FracPhys Maps for training.
  \item \textbf{FracGen}, a video generation model that jointly predicts fracture RGB video and physical fields, trained with physics-informed losses encouraging consistency between physics maps.
  \item A benchmark with new evaluation metrics for evaluating fracture video generation across diverse materials, applied
  forces, and fracture behaviors.
  \item Extensive experiments showing FracGen generates physically plausible, controllable
  fracture that generalizes across unseen objects, outperforming existing baselines
  both visually and physically.
\end{enumerate}

\vspace{-0.3cm}
\section{Related Work}
\label{sec:related}
\paragraph{Physics-informed 3D Simulation.}
A recent line of work grounds 3D dynamics in continuum mechanics. PhysGaussian ~\citep{Xie_2024_CVPR} couples the Material Point Method (MPM) \citep{hu2018moving} with 3D Gaussians, treating each Gaussian as both a rendering primitive and physical particle, while Gaussian Splashing \citep{Feng_2025_CVPR} adopts Position-Based Dynamics \citep{macklin2016xpbd} for cohesive solid-fluid simulation. However, these solvers require manual scene specification and tedious parameter tuning. To tackle this, PhysDreamer \citep{zhang2024physdreamer} optimizes MPM material parameters against generative reference videos using differentiable simulation, whereas DreamPhysics \citep{10.1609/aaai.v39i4.32389} optimizes directly via Score Distillation Sampling (SDS) \citep{poole2023dreamfusion}. Physics3D \citep{liu2024physics3d} and OmniPhysGS \citep{lin2025omniphysgs} further expand SDS-driven optimization to richer constitutive models, incorporating viscoelastic damping or multi-material mixtures (elastic, plastic, fluid) within a single scene.
\vspace{-0.2cm}
\paragraph{Physics-informed Video Generation.}
With advances in video generation models, recent works seek to inject faithful physical realism through two main strategies. The first conditions generation on
explicit control signals: Force Prompting~\citep{gillman2026force} fine-tunes a video diffusion model on synthetic force-annotated data, PhysCtrl \citep{physctrl2025} controls video diffusion using predicted 3D point-trajectories conditioned on forces and material traits, and PhyCo \citep{narayanan2026phyco} enables pixel-aligned ControlNet guidance over physical fields (friction, deformation, force) refined by VLM reward optimization. The second strategy enforces physics via high-level reasoning and reward alignment: DiffPhy~\citep{zhang2025thinkdiffusellmsguidedphysicsaware} uses LLM context-builder alongside a MLLM physical critic, while PhyGDPO~\citep{cai2025phygdpo} post-trains a video models using physics-aware groupwise preference optimization. However, these methods focus on rigid-body motion or simple deformation, lacking mechanisms for material failure. To the best of our knowledge, FracGen is the first model to generate complex fracture dynamics across the full deformation-to-fracture progression.

\section{Background}
\label{sec:preliminaries}
\paragraph{Gaussian--MPM Simulation.}
\looseness=-1 3D Gaussian Splatting (3DGS) \citep{kerbl3Dgaussians} represents a scene using anisotropic Gaussian kernels that carry geometry and appearance. PhysGaussian \citep{Xie_2024_CVPR} couples this representation with the Material Point Method (MPM) \citep{hu2018moving}, associating particles with evolving positions $x_p^t$, velocities $v_p^t$, and deformation gradients $\bF_p^t$. MPM updates these states through transfers between particles and a background grid, while the Gaussians deform with the material to render its evolving appearance.
The deformation map $\phi$ transports a material point from its reference position $x_p$ to $x_p^t=\phi(x_p,t)$. Its gradient, $\bF_p^t=\nabla_{x_p}\phi(x_p,t)$, encodes local stretching, compression, rotation, and shearing. The polar decomposition $\bF_p=\bm R_p\bm S_p$ separates rigid rotation $\bm R_p$ from stretch $\bm S_p$, allowing us to distinguish shape change from rigid motion.

\paragraph{Constitutive Models.}
Strain measures local deformation relative to the reference configuration, while stress describes the internal forces per unit area transmitted through the material. A constitutive model relates deformation to stress. For hyperelastic materials, a strain energy density $\psi(\bF)$ determines the stress response, which can be nonlinear while remaining reversible. In isotropic linear elasticity, Hooke's law gives $\boldsymbol{\sigma}=\lambda_{\mathrm{L}}\operatorname{tr}(\boldsymbol{\varepsilon})\mathbf{I}+2\mu\boldsymbol{\varepsilon}$, where $\lambda_{\mathrm{L}}$ and $\mu$ are the Lamé parameters. FracSim computes stress from the material's strain energy and introduces damage-dependent degradation, while our generator uses the linear-elastic relationship to motivate an approximate consistency loss in latent space.


\paragraph{Particle State Update with MPM.} For particle $p$ at simulation time $t$, let $x_p^t$, $v_p^t$, and $\bF_p^t$ denote its position, velocity, and deformation gradient, respectively. At each step, MPM transfers particle quantities to a background Eulerian grid, solves the momentum equation, and transfers the results back to update these states. Since each particle carries both appearance and physical state, a single representation supports simulation and rendering. Under a first-order approximation of $\phi$, each Gaussian's center moves from its reference position $x_p$ to $x_p^t$, and its reference covariance $A_p$ becomes $A_p^t=\bF_p^t A_p(\bF_p^t)^\top$. Its spherical harmonic (SH) appearance rotates with the particle rotation $\bm R_p$, while its opacity $o_p$ remains fixed.

\paragraph{Conditional Video Generation}
\label{sec:control_vid_gen}
We build on a latent video generator \citep{wan2025wan} that encodes an RGB video $\mathbf V$ as $z_1=\mathcal E(\mathbf V)$ using a 3D VAE. Under flow matching \citep{lipman2023flow}, a transformer $v_\theta$ learns to transport Gaussian noise $z_0\sim\mathcal N(0,\mathbf I)$ to video latents along the linear path $z_t=t z_1+(1-t)z_0$. The training objective is
\begin{equation}
\mathcal L_{\mathrm{FM}}
=
\mathbb E_{t,z_0,z_1,c}
\left[
\left|
v_\theta(z_t,t,c)-(z_1-z_0)
\right|^2
\right],
\label{eq:fm_obj}
\end{equation}
where $t\sim\mathcal U(0,1)$ and $c$ denotes conditioning inputs. We extend this formulation to jointly generate fracture videos and physical maps under specified material and force conditions.

\section{FracSim and FracGen}
\label{sec:main_method}
We discuss in \S\ref{subsec:FracSim} how we design \textbf{FracSim}, which outputs dynamic fracture physics maps (FracPhys) alongside fracture videos, forming corpus for training \textbf{FracGen}, as detailed in \S\ref{subsec:frac_gen}.

\begin{figure}[t]
    \centering
    \includegraphics[width=\linewidth, trim={0cm 2.7cm 0cm 2.8cm}, clip]{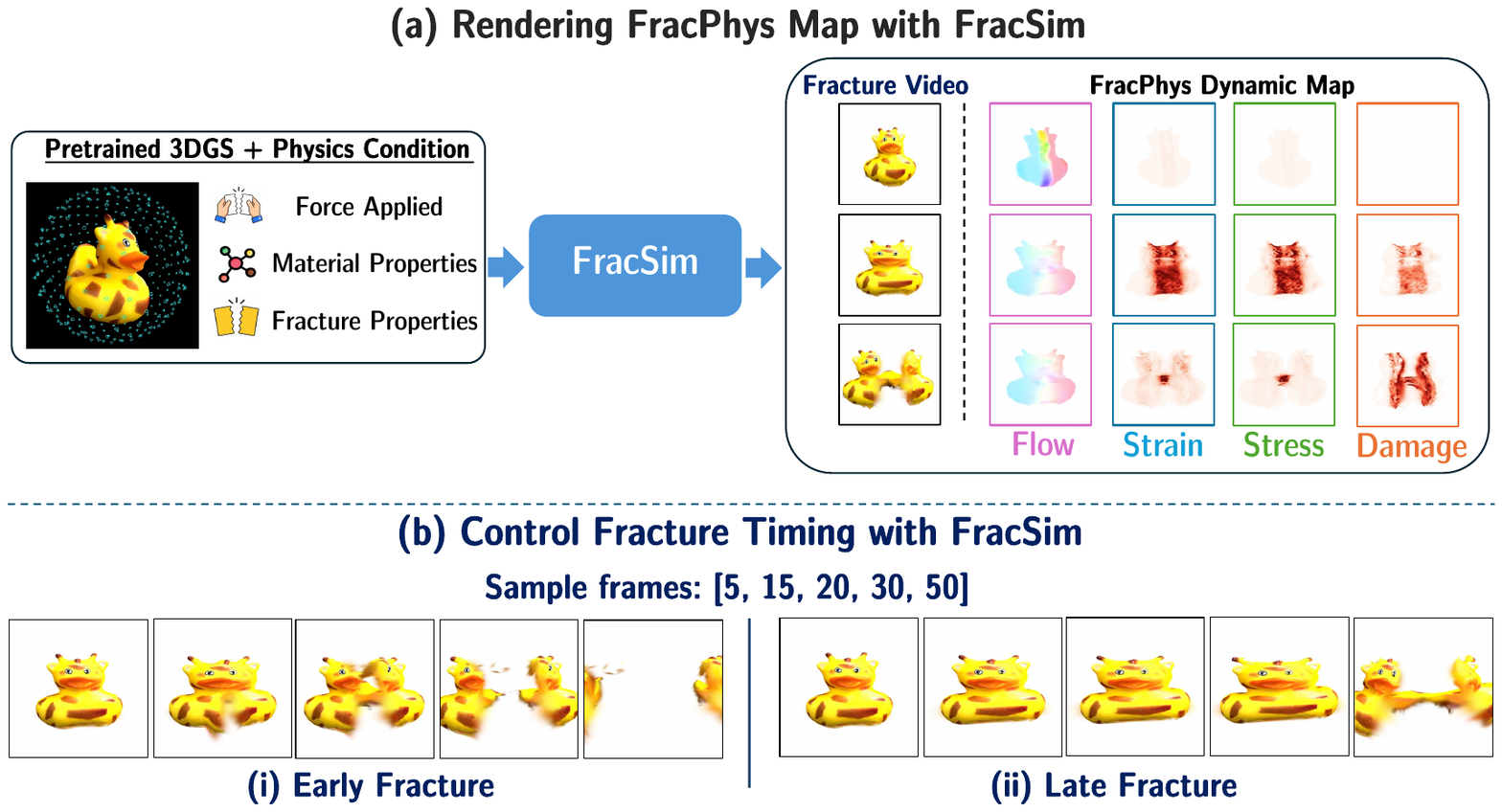}
    \caption{FracSim generates physically plausible fracture dynamics spanning deformation to structural failure. (a) Alongside RGB video, FracSim outputs FracPhys map. (b) Our damage model enables explicit control over fracture timing under fixed forces and material properties.}
    \label{fig:FracSim}
\end{figure}

\subsection{FracSim: Simulating Fracture Video and Rendering FracPhys Maps}
\label{subsec:FracSim}
We integrate a continuum damage model into MPM to generate dynamic physics maps alongside RGB frames. We refer to these maps as FracPhys maps and discuss how to extract them below.

\textbf{Flow.} Tensile load is prescribed via particle velocities (referred as flow), making $\bm{v}_p \in \mathbb{R}^{3}$ the most direct kinematic signal in the simulation. Rather than estimating this field from rendered frame via optical flow, we directly extract it from the MPM state to preserve exact direction and magnitude.

\textbf{Strain.} We require a strain measure that vanishes under rigid rotation yet remains meaningful at large deformation. The Green-Lagrange strain tensor satisfies both requirements:
\begin{equation}
\boldsymbol{\varepsilon}_p = \tfrac{1}{2}\left(\bF_p^{\top} \bF_p - \Id\right)
= \tfrac{1}{2}\left(\bm{S}_p^{\top} \bm{S}_p - \Id\right),
\label{eq:green_strain_particle}
\end{equation}
where the polar decomposition $\bF_p = \bm{R}_p\bm{S}_p$ isolates stretch $\bm{S}_p$ from rotation $\bm{R}_p$, which cancels exactly. The result is rotation-invariant, zero at rest, and nonzero only when genuine deformation occurs. We record its Frobenius norm $\|\boldsymbol{\varepsilon}_p\|_F$ as a scalar per particle.


\textbf{Stress.} We compute the Kirchhoff stress tensor from the strain energy density $\psi(\bF_p)$:
\vspace{-2pt}
\begin{equation}
\boldsymbol{\tau}_p
=
\frac{\partial\psi}{\partial\bF_p}\bF_p^\top.
\end{equation}
\looseness=-1 Damage degrades this tensor before it is used in the MPM momentum update. We then convert the degraded tensor to Cauchy stress and extract its von Mises magnitude for rendering, as described below.

\textbf{Damage.}
We augment PhysGaussian with a scalar damage variable $d_p\in[0,1]$, where $0$ denotes intact material and $1$ denotes complete failure. To represent this degradation at the particle level, we use a kinematic proxy based on the maximum principal stretch $\lambda_{\max}(\bF_p)$, the largest singular value of $\bF_p$, following \citep{Patnaik2021}:
\begin{equation}
d^t_{p,\mathrm{raw}}
=
\operatorname{clamp}\!\left(
\frac{\lambda_{\max}(\bF_p^t)-\lambda_{\mathrm{onset}}}
{\lambda_{\mathrm{crit}}-\lambda_{\mathrm{onset}}},
0,1\right),
\qquad
d_p^t
=
\max\!\left(d_p^{t-1},d^t_{p,\mathrm{raw}}\right).
\label{eq:damage}
\end{equation}
We initialize $d_p^0=0$ and require
$\lambda_{\mathrm{crit}}>\lambda_{\mathrm{onset}}\geq1$.
Under increasing stretch, damage begins at $\lambda_{\mathrm{onset}}$ and grows linearly to complete failure at $\lambda_{\mathrm{crit}}$. The running maximum preserves accumulated damage during unloading, enforcing irreversibility. Adjusting these thresholds controls fracture timing under fixed loading and material parameters (Fig.~\ref{fig:FracSim}b).

Damage feeds back into the simulation by degrading the stress tensor:
\begin{equation}
\tilde{\boldsymbol{\tau}}_p
=
(1-d_p)\boldsymbol{\tau}_p.
\label{eq:degraded_stress}
\end{equation}
This degraded tensor is used in the MPM momentum update. As $d_p$ approaches one, the particle progressively loses its ability to sustain load, allowing separation. For the stress map, we convert the degraded Kirchhoff stress to Cauchy stress,
$\tilde{\boldsymbol{\sigma}}_p=\tilde{\boldsymbol{\tau}}_p/J_p$,
where $J_p=\det(\bF_p)$, and compute its equivalently degraded von Mises scalar $\tilde{\boldsymbol{\sigma}}_{\mathrm{vM},p}$.
Damage is the sole softening mechanism; we do not model plastic flow (more discussion in \S\ref{subsecsion:sim_train_details}).

\paragraph{Rendering FracPhys Maps.} 
As discussed in \S\ref{sec:preliminaries}, 3DGS composites pixel color using appearance term
$\mathrm{SH}(l_p;\mathcal{C}_p)$. We replace this with any per-particle quantity $q_p$:
\vspace{-5pt}
\begin{equation} M_q = \sum_{p \in \mathcal{P}} \alpha_p\, q_p \prod_{j=1}^{p-1}(1-\alpha_j). 
\label{eq:physmap_rendering} 
\vspace{-5pt}
\end{equation} 
where $q_p = \tilde{\bsig}_{\text{vM},p}$ for stress, $\|\boldsymbol{\varepsilon}_p\|_F$ for
strain, $\bm{d}_p$ for damage, or $\bm{v}_p \in \mathbb{R}^3$ for velocity. For vector-valued
$q_p$, compositing applies component-wise. For every RGB frame, this yields four spatially
and temporally aligned \textbf{FracPhys Maps} visualizing the full evolution of fracture
dynamics.
\begin{figure}[t]
    \centering
    \includegraphics[width=\linewidth, trim={0cm 2.5cm 0cm 5.5cm}, clip]{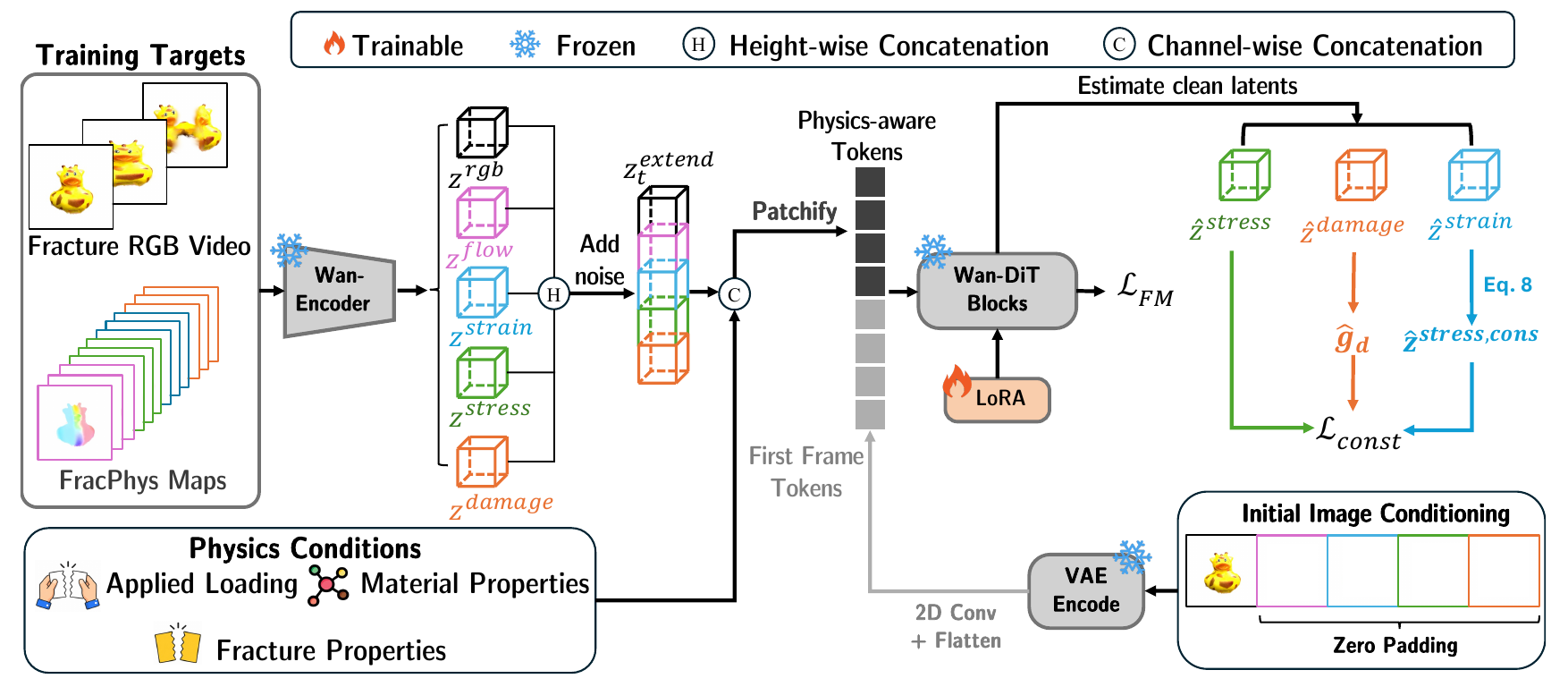}
\vspace{-10pt}
\caption{\textbf{FracGen overview.} RGB fracture videos and FracPhys Maps are encoded by a frozen VAE and concatenated along the spatial height dimension for joint generation. Applied loading, material properties and fracture properties provide channel-wise conditioning. The initial object image supplies reference tokens. We adapt the transformer with LoRA using flow-matching supervision and a physics-informed latent consistency regularizer coupling strain, stress, and damage representations.}
    \label{fig:fracgen}
\end{figure}

\subsection{FracGen: Teaching Video Generator Fracture Dynamics}
\label{subsec:frac_gen}
Given an object image along with physics controls (tensile load, material, and fracture properties), we describe how we train FracGen to jointly predict fracture RGB video alongside its corresponding dynamic physics maps $\hat{\mathbf{V}} = \{\mathbf{\hat{V}}^\text{rgb}, \mathbf{\hat{V}}^m\}$. This joint prediction enhances both visual quality and physics accuracy, enabling inspection \textbf{without requiring any external physics simulators}.

\paragraph{Constructing Joint Latents.} We adopt the latent flow-matching video generation model \S\ref{sec:control_vid_gen}. To support joint prediction, we construct an extended clean latent $z^\text{extend}$, which is formed by height concatenating encoded version of video latents and its dynamic physics map $z^m$ with $K=4$:
\begin{equation}
z^{\text{extend}} = \mathrm{Concat}_{H}\big(z^\text{rgb}, z^\text{flow}, z^\text{strain}, z^\text{stress},z^\text{damage}\big) \in \mathbb{R}^{(1+T') \times (K{+}1)H' \times W' \times C}.
\end{equation}
Following \citep{chen20254dnex}, this design choice serves as an effective alternative to channel-wise concatenation without adding parameters. Preprocessed physics maps are projected into latent space via a frozen pretrained 3D VAE encoder $z^{m} = \mathcal{E}(\mathbf{V}^m)$, avoiding costly retraining while preserving reconstruction fidelity (see \S\ref{sec:Appendix} for more details). Finally, noise is injected into $z^\text{extend}$ to form $z_t^\text{extend}$.

\paragraph{Conditioning Physics Signals.} FracGen is conditioned on three physics-guided signals: applied forces, material properties, and fracture properties. We detail their latent space encoding below: 
\begin{enumerate}[leftmargin=1.8em]
    \item \textbf{Applied Loading.} We use opposing velocity signals mimicking a universal testing machine (UTM) and render it as moving Gaussian blobs across frames. Blob centers denote grip locations, fixed isotropic covariances define spatial extent, and displacements reflect velocity vectors. We use a frozen 3D VAE to encode this dynamic signal into latent $y_{\text{force}} \in \mathbb{R}^{(1+T') \times H' \times W' \times 16}$, spatially and temporally aligned with the RGB video representation.

    \item \textbf{Material Properties.} Material deformation and fracture depend heavily on Young's modulus $E$ and Poisson's ratio $\nu$. Instead of using an explicit encoder, we construct $y_{\text{mat}} \in \mathbb{R}^{(1+T') \times H' \times W' \times 2}$ by broadcasting normalized parameter maps across space and time. We normalize $E \in [10^3, 10^9]$ Pa as $\hat{E} = \frac{\log(E) - 6}{3}\in [-1, 1]$ and $\nu \in [0.1, 0.5]$ as $\hat{\nu} = \frac{\nu - 0.3}{0.2}\in [-1, 1]$.
    
    \item \textbf{Fracture Properties.} To control fracture behavior, we first define fracture categories using a bipolar one-hot category code $\mathbf{c} \in \{-1,+1\}^2$ (low-stretch fracture vs. high-stretch fracture). For fracture conditions, we define the fracture onset and critical stretch, at which fracture initiates and fully fails, respectively, as discussed in \S\ref{subsec:FracSim} via $\lambda' = [\lambda_{\text{onset}}, \lambda_{\text{crit}}]$. We normalize $\lambda \in [1, 10]^2$ to $[-1, 1]^2$ via $\frac{\lambda' - 5.5}{4.5}$ and broadcast these across space and time via $y_{\text{frac}} \in \mathbb{R}^{(1+T') \times H' \times W' \times 4}$.
    \end{enumerate}

Following Wan 2.1-1.3B-Control \citep{wan2025wan}, we allocate a 32-channel conditioning layout: 16 channels for $y_{\text{force}}$; 6 channels for $y_{\text{frac}}$ and $y_{\text{mat}}$, and 10 zero-padded channels. We concatenate $z_t^{\text{extend}}$ with these physics-condition channels along the channel axis before patchify as shown in Fig.~\ref{fig:fracgen}.

\paragraph{Conditioning Intact Object Image.} We anchor the generated video's first frame to the intact object, allowing the model to infer shape, texture, and geometry priors before force application. Following Wan-DiT's \citep{wan2025wan}, we encode this single RGB image using the same frozen 3D VAE. To align with the height of the multi-modal extended latent $z_t^{\text{extend}}$, we zero-pad the reference latent along the spatial height axis to indicate the absence of initial physics-map references. The padded reference latent is then processed by a 2D convolution matching the DiT patch size and flattened into reference tokens, which are prepended to the physics-aware input tokens as shown in Fig.\ref{fig:fracgen}.
\begin{figure}[t!]
    \centering
    \begin{subfigure}{\linewidth}
        \centering
        \includegraphics[width=\linewidth, trim={1.2cm 5cm 0.6cm 3cm}, clip]{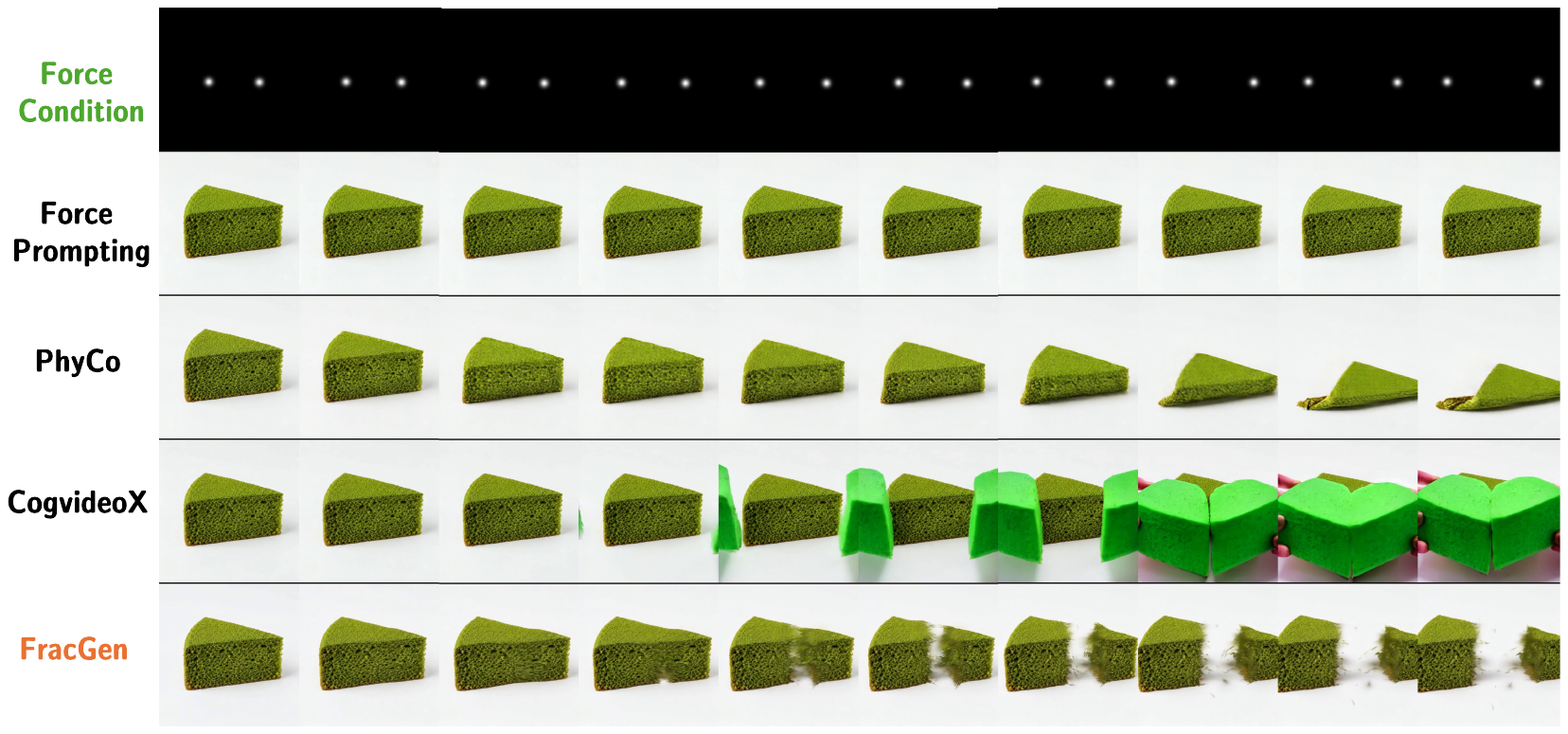}
        \caption{Cake tearing with early break.}
        \label{fig:main_qual_part1}
    \end{subfigure}
    
    \vspace{0.1em} 
    
    \begin{subfigure}{\linewidth}
        \centering
        \includegraphics[width=\linewidth, trim={1.2cm 5cm 0.6cm 3cm}, clip]{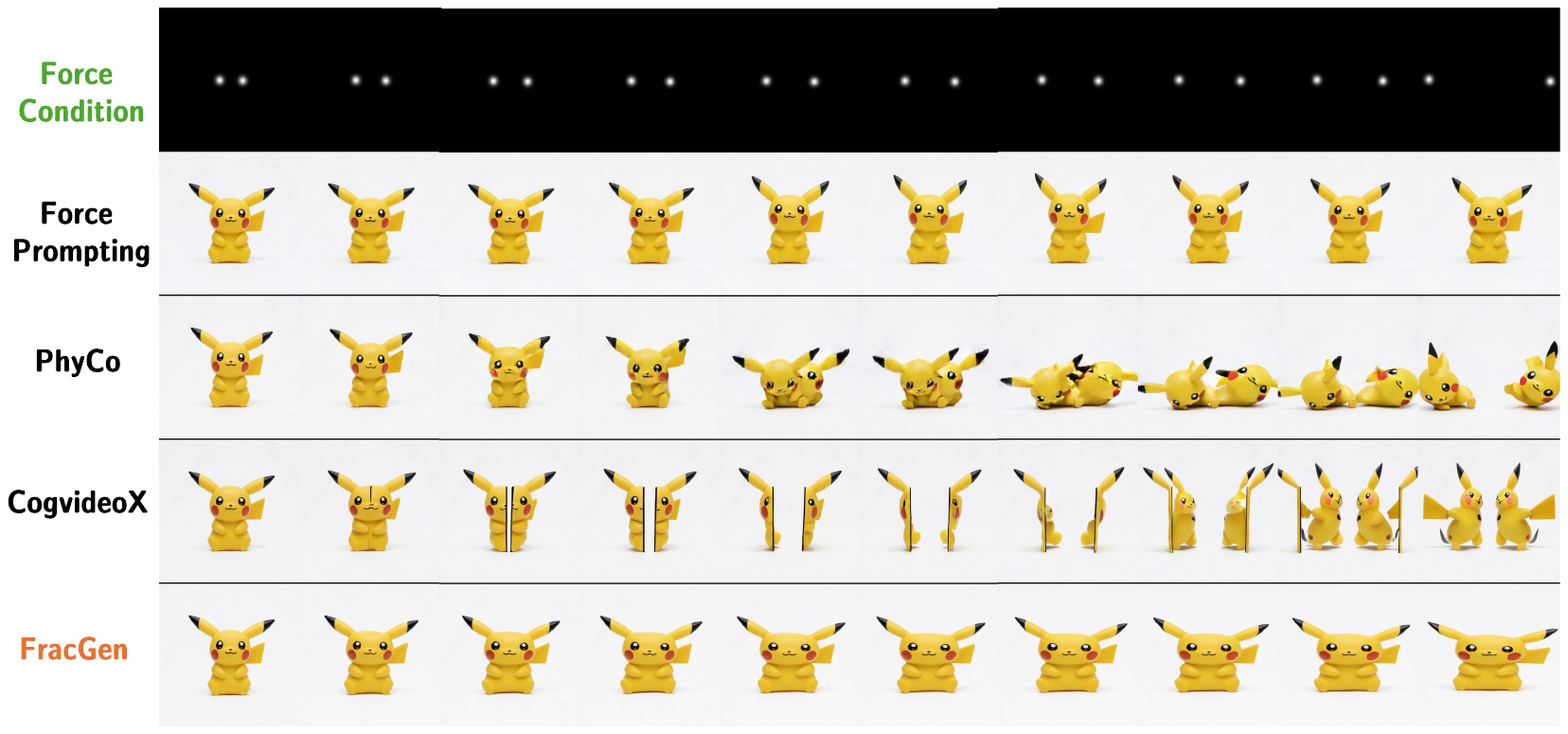}
        \caption{Pikachu extreme stretching.}
        \label{fig:main_qual_part2}
    \end{subfigure}
    \vspace{-15pt}
    \caption{\textbf{Qualitative comparison on unseen objects.} We compare generated responses to pulling conditions for (a) a cake that tears after limited stretching and (b) a toy that undergoes sustained elongation. FracGen captures these distinct behaviors while better preserving object appearance; competing methods show limited deformation, collapse or changes in object identity. The top row of each panel shows the force conditioning, and frames progress from left to right.}
    \label{fig:main_qual}
    \vspace{-15pt}
\end{figure}

\paragraph{Training Losses.}
 \looseness=-1 Alongside the latent flow-matching loss (\S\ref{sec:control_vid_gen}), we introduce a \textbf{latent constitutive regularizer} inspired by linear elasticity and damage-induced stress degradation. Our physical maps encode scalar strain magnitude, von Mises stress and damage, which are further compressed by a pretrained VAE. We therefore learn an approximate relationship among their latent representations.

During training, we estimate the clean joint latent from the noisy input and the model's predicted flow using the training scheduler. We extract the corresponding strain, stress, and damage latents, denoted by $\hat{z}^{\mathrm{strain}}$, $\hat{z}^{\mathrm{stress}}$, and $\hat{z}^{\mathrm{damage}}$. From the damage latent, we construct a channel-wise gate
$\hat{g}_d=\operatorname{sigmoid}(\hat{z}^{\mathrm{damage}})$,
with the same shape as the stress latent. Motivated by the isotropic linear-elastic relationship
$\boldsymbol{\sigma}=\lambda_{\mathrm{L}}\operatorname{tr}(\boldsymbol{\varepsilon})\mathbf{I}+2\mu\boldsymbol{\varepsilon}$,
we define an approximate latent stress target:
\begin{equation}
\hat{z}^{\mathrm{stress,cons}}
=
\lambda_{\mathrm{L}} f_{\eta}(\hat{z}^{\mathrm{strain}})
+ 2\mu g_{\xi}(\hat{z}^{\mathrm{strain}}),
\end{equation}
where $f_{\eta}$ and $g_{\xi}$ are zero-initialized $1\times1\times1$ convolutions that learn mappings across latent channels. The Lamé parameters are computed from each sample's Young's modulus $E$ and Poisson's ratio $\nu$:
\begin{equation}
\small
\lambda_{\mathrm{L}}
= \frac{E\nu}{(1+\nu)(1-2\nu)},
\qquad
\mu = \frac{E}{2(1+\nu)}.
\end{equation}
The regularizer couples the predicted stress latent to this target through the damage gate:
\begin{equation}
\mathcal{L}_{\mathrm{const}}
=
w(t)
\left|
\hat{z}^{\mathrm{stress}}
-
(1-\hat{g}_d)\odot
\hat{z}^{\mathrm{stress,cons}}
\right|_2^2,
\label{eq:const_loss}
\end{equation}
where $\odot$ denotes element-wise multiplication and $w(t)$ assigns greater weight at lower noise levels. When the gate approaches zero, the stress latent is encouraged to match the constitutive approximation; when it approaches one, the target approaches zero latent, a soft degradation prior rather than exact zero stress. This provides a latent-space analogue of stress degradation, without assuming that the VAE coordinates directly represent physical stress or damage. The final training objective is
\begin{align} 
\mathcal{L} = \mathcal{L}_{\mathrm{FM}} + \lambda_{\mathrm{const}} \cdot \mathcal{L}_{\mathrm{const}}, 
\label{eq:final_loss} 
\end{align} 
where $\lambda_{\mathrm{const}}$ controls the regularization weight. We train FracGen in three stages. First, we train on RGB fracture videos with physics conditioning. Second, we initialize from this checkpoint and introduce joint prediction of RGB videos and FracPhys Maps. Third, we fine-tune the model with the complete objective in Eq.~\ref{eq:final_loss}.
\section{Experiments}
\label{sec:experiments}
\paragraph{Simulation and Training Details.} For simulation, we sample 10 objects from the Objaverse dataset \citep{objaverse} and SketchFab platform, covering two common material classes in stretch-to-tear fracture: 5 high-stretch objects (rubber toys) and 5 low-stretch objects (croissant, cake, muffin, bread, and bread roll). We then use FracSim to yield a paired dataset of RGB fracture videos and dynamic physics maps for generative training by varying different force condition, material properties, and fracture properties (more details can be seen in \S\ref{sec:Appendix}). For training, we leverage this simulated data to train our FracGen model. Among 10 objects, simulated results for 6 objects (3 low-stretch and 3 high-stretch objects) with varying conditions are used for training, yielding 4,212 training samples in total. More details on simulation and training setting could be found in \S\ref{sec:Appendix}).

\begin{figure}[t]
    \centering
    \includegraphics[width=1.02\linewidth, trim={0cm 9cm 0cm 3.2cm},clip]{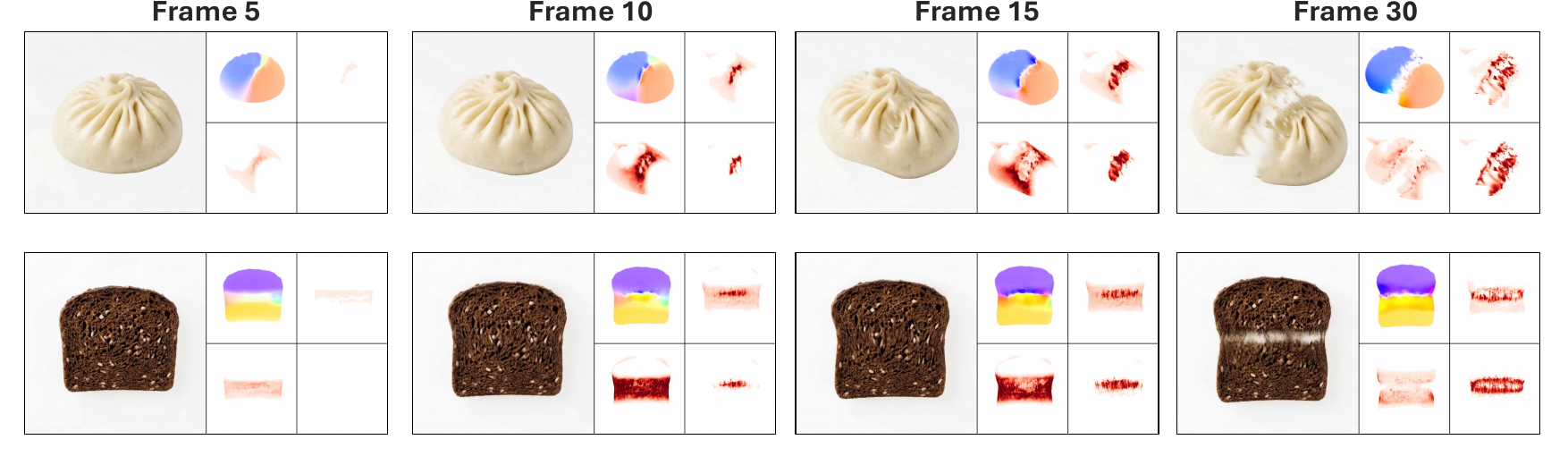}
\vspace{-20pt}
    \caption{\textbf{Generalization to in-the-wild objects}. FracGen only needs one single object image and jointly generates the fracture video (left panel) and its FracPhys maps (right panel, 2×2 grid in top-left to bottom-right order: flow, strain, stress, damage), under varying force directions.}
    \label{fig:phys_map_pred}
\end{figure}

\paragraph{Benchmark.} We hold out 4 of the 10 objects for testing against groundtruth, which is mainly for quantitative evaluation against groundtruth. Note that these evaluation set includes 4 objects with varying physics conditions such as forces, material properties, and fracture properties, yielding 2,808 testing samples in total (single view images for these objects are shown in Fig.\ref{fig:app_10_objects}). In addition, we collect another set of single object images (see Fig.\ref{fig:app_10_random_objects}) with no 3DGS reconstruction, and hence no simulated video groundtruth. This is to qualitatively show real-world scenario where a user want to fracture an object from a single photo. These are used to demonstrate generalization to new objects, shown in Fig.\ref{fig:main_qual_part1}, Fig.\ref{fig:main_qual_part2}, Fig.\ref{fig:phys_map_pred}. We provide more results on our project website.

\begin{table}[t]
\centering
\small
\setlength{\tabcolsep}{2pt}
\caption{Quantitative results on held-out test set. All baselines are fine-tuned on the same dataset.}
\vspace{-5pt}
\begin{tabular}{@{}lccccc@{}}
\toprule
\textbf{Method} & \textbf{Physics Map Pred} & \textbf{FVD} $\downarrow$ & \textbf{LPIPS} $\downarrow$ & \textbf{PSNR} $\uparrow$ & \textbf{SAC} $\uparrow$ \\
\midrule
ForcePrompting-FT \citep{gillman2026force} & \textcolor{red}{\(\times\)} & 933.21 & 0.29 & 15.93 & 0.092 \\
PhyCo-FT \citep{narayanan2026phyco} & \textcolor{red}{\(\times\)} & 832.12 & 0.33 & 14.18 & 0.434 \\
CogVideoX-5B-I2V-FT \citep{yang2025cogvideox} & \textcolor{red}{\(\times\)} & 1533.21 & 0.32 & 14.12 & 0.432 \\
\midrule
\textbf{FracGen} & \textcolor{green!60!black}{\(\checkmark\)} & \textbf{266.17} & \textbf{0.15} & \textbf{21.19} & \textbf{0.946} \\
\bottomrule
\end{tabular}
\label{tab:main_quan}
\end{table}

\paragraph{Evaluation Metrics.}
We evaluate visual quality using standard metrics: FVD \citep{unterthiner2019fvd}, LPIPS \citep{zhang2018perceptual}, and PSNR. Since existing physics metrics such as VideoPhy \citep{bansal2024videophyevaluatingphysicalcommonsense}, VideoPhy2 \citep{bansal2026videophy}, and PhysicsIQ \citep{motamed2026generative} focus on general commonsense rather than fracture, we introduce several fracture-aware metrics:

\begin{enumerate}[leftmargin=1.8em]
    \item \textbf{Stretching Axis Coherence (SAC).} SAC measures whether the object's dominant
    deformation direction aligns with the ground-truth stretching axis. We estimate the
    dominant stretching direction from point tracks using \citep{lai2026a} and compute its alignment with the
    ground-truth axis; higher scores indicate better alignment (see \S\ref{sec:Appendix} for more details).
    
    \item \textbf{Damage Progression Alignment (DPA).} Because FracGen conditions on fracture timing and predicts damage maps, we evaluate damage accumulation schedule. We primarily used this for ablation studies since baselines lack this capability, further details are in the \S\ref{sec:Appendix}.

    \item \textbf{Constitutive Consistency ($\mathbf{R}^2_{cons}$).} Unlike other metrics that evaluates physics map individually, this one evaluate the consistency among physics map. Specifically, we use the damage map to mask out already-failed regions and measure linear elasticity in the material that remains intact, precisely the regime where Hooke's law applies. The metric requires no ground truth, as it tests whether the model's own stress and strain predictions are mutually consistent. Like DPA, we use it for ablation studies rather than baseline comparisons. Further details are in the \S\ref{sec:Appendix}.
\end{enumerate}

\vspace{-8pt}
\paragraph{Baselines.} We compare FracGen against recent physics-aware video generation models (e.g., ForcePrompting, PhyCo) and video generation baselines (e.g., CogVideoX). We do not compare against PhysCtrl~\citep{physctrl2025}, as it encodes force as a single force vector, which cannot represent our dual opposing forces. In contrast, ForcePrompting and PhyCo encode force as rendered, pixel-aligned control videos, so we extend them to our two-grip setup by rendering both grips in the control signal, and fine-tune them on the same data (more details in \S\ref{sec:Appendix}). As shown in Tab.~\ref{tab:main_quan} and Fig.~\ref{fig:main_qual}. FracGen achieves superior performance across all metrics, excelling in both video quality and physical alignment (SAC). Qualitatively, baselines fail to capture realistic stretching, and fracture progressions, while FracGen faithfully captures this behavior. More results are on our project website.

\begin{figure}[t]
    \centering
    \includegraphics[width=\linewidth, trim={0.3cm 2.3cm 0.3cm 2.5cm},clip]{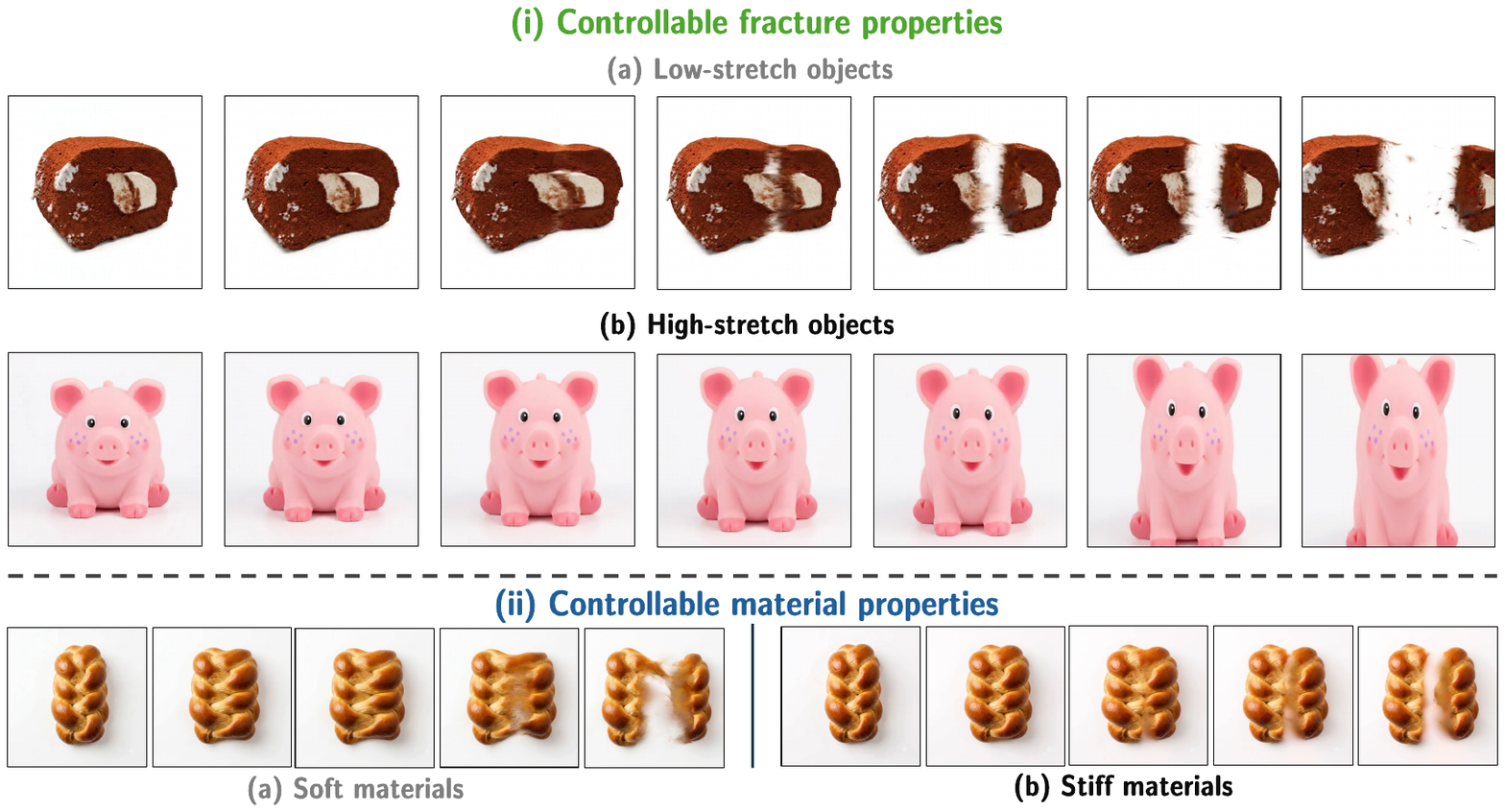}
\vspace{-20pt}
    \caption{\textbf{Controlled deformation and tearing.} (i) Generated sequences for objects with low- and high-stretch fracture settings. (ii) Changing material stiffness for the same object produces different deformation and tearing behavior.}
    \label{fig:more_qual}
\vspace{-10pt}
\end{figure}

\paragraph{Controllable FracPhys Map and Fracture Prediction.} As shown in Fig. \ref{fig:phys_map_pred}, FracGen generates physics maps that align well with fracture videos under varying force conditions (vertical and diagonal tearing directions). Furthermore, Fig. \ref{fig:more_qual} demonstrates enhanced controllability by conditioning on specific material and fracture properties across different settings: low-stretch objects (featuring low stiffness and low critical damage thresholds) versus high-stretch objects (featuring high stiffness and high thresholds). Additional video results are available on the supplement website.

\begin{wraptable}{r}{0.5\textwidth}
\vspace{-1.25em}
\centering
\small
\setlength{\tabcolsep}{1pt}
\caption{\looseness=-1 Ablation: \textcolor{orange}{\textbf{video quality}}, \textcolor{blue}{\textbf{physics alignment}}, and evaluation of FracPhys map prediction (MSE). Stage 1 has no physics map; hence \textcolor{blue}{\textbf{DPA}} and FracPhysMap evaluation are not applicable.}
\vspace{-5pt}
\begin{tabular}{@{}lccc@{}}
\toprule
\textbf{Metric} & \textbf{Stage 1} & \textbf{Stage 2} & \textbf{Stage 3} \\
& (\textcolor{red}{\(\times\)} FracPhys) & (\textcolor{green!60!black}{\(\checkmark\)}, w/o Eq.\ref{eq:const_loss}) & (\textcolor{green!60!black}{\(\checkmark\)}, w. Eq.\ref{eq:const_loss}) \\
\midrule
\rowcolor{gray!20}
\multicolumn{4}{c}{\textit{\textbf{Video Quality \& Physics Alignment}}} \\
\textcolor{orange}{\textbf{FVD} $\downarrow$} & 298.04 & 289.65 & \textbf{266.17} \\
\textcolor{orange}{\textbf{LPIPS} $\downarrow$} & 0.1555 & 0.1550 & \textbf{0.1502} \\
\textcolor{orange}{\textbf{PSNR} $\uparrow$} & 20.73 & 20.82 & \textbf{21.19} \\
\textcolor{blue}{\textbf{SAC} $\uparrow$} & 0.9460 & 0.9393 & \textbf{0.9462} \\
\textcolor{blue}{\textbf{DPA} $\uparrow$} & -- & 0.438 & \textbf{0.536} \\
\midrule
\rowcolor{gray!20}
\multicolumn{4}{c}{\textit{\textbf{FracPhys Map Evaluation}}} \\
\textbf{Flow} $\downarrow$ & -- & 0.03735 & \textbf{0.01147} \\
\textbf{Stress} $\downarrow$ & -- & 0.01464 & \textbf{0.00429} \\
\textbf{Strain} $\downarrow$ & -- & \textbf{0.02017} & 0.02681 \\
\textbf{Damage} $\downarrow$ & -- & 0.02904 & \textbf{0.02740} \\
\textbf{Overall} $\downarrow$ & -- & 0.02530 & \textbf{0.01749} \\
$\mathbf{R}^2_{Cons}$ $\uparrow$ & -- & 0.378 & \textbf{0.902} \\
\bottomrule
\end{tabular}
\label{tab:combined_ablation}
\vspace{-25pt}
\end{wraptable}

\paragraph{Ablation: Effects of Joint Prediction.} Stage 1 (physics conditioning, RGB only) serves as our base model. Adding joint FracPhys prediction (Stage 2) improves video quality (FVD $298\!\rightarrow\!290$) and additionally yields pixel-aligned physical fields that allow the generated dynamics to be inspected directly (Tab.~\ref{tab:combined_ablation}).

\paragraph{Ablation: Effects of Constitutive Loss.} With constitutive loss, Stage 3 yields the largest gains across all levels. Video quality improves (FVD $290\!\rightarrow\!266$), and damage progression aligns more closely with ground truth (DPA $0.438\!\rightarrow\!0.536$). Predicted maps become more accurate: overall MSE drops from $0.0253$ to $0.0175$, while strain error increases slightly, which we attribute to the loss prioritizing stress--strain coupling over per-map fidelity. We also observe a rise in consistency ($R^2_{\text{cons}}$ $0.378\!\rightarrow\!0.902$). As reference, the $\mathbf{R}^2_{Cons}$ score for groundtruth physics maps is $\textbf{0.948}$.

\section{Discussion}
\label{sec:conclusion}
\looseness=-1 We show that pretrained video generators can be adapted to learn material-dependent stretching and tearing from simulation-derived supervision. To our knowledge, FracGen is the first video generation framework to jointly predict material-dependent tearing and the associated physical fields from a single image and specified physical conditions. FracSim provides paired videos and physical maps for joint prediction, while physics-consistency losses encourage agreement among the predicted fields. These allow us to inspect how estimated strain, stress and damage evolve. As video models improve and the materials community releases more datasets, we envision a feedback loop: experimental data improves generative models, and fast predictions help prioritize new experiments. With experimental validation, this could accelerate material-response prediction and support materials discovery.

Our work focuses on deformation and tearing under tensile loading. Brittle fracture and fragmentation under impact are outside our training and evaluation scope. Extending FracGen to these material types would require appropriate simulation data and failure models. The model also inherits the assumptions of its training simulator, so any errors in FracSim would also reflect in FracGen. Finally, multi-step diffusion sampling limits inference speed. Few-step distillation could reduce this cost, provided it preserves fracture behavior and consistency among the predicted physical fields.

\subsection*{AI use statement}
We used generative AI tools to assist with language editing and translation, drafting parts of the paper text, summarizing existing literature, proposing title and keyword candidates, setting up environments for baseline methods, cleaning and reformatting datasets, and building the project website. We did not use generative AI to generate synthetic datasets, propose or refine hypotheses, design research methodology or experiments, or implement our method; theoretical and proof-related uses are not applicable to this work. All AI-assisted outputs were reviewed by the authors: drafted and translated text was edited, baseline and data scripts were verified by running them and inspecting outputs, and literature summaries were checked against the original papers. We take responsibility for the final content of this work, including text, claims or artifacts produced with the aid of generative AI.




\bibliography{iclr2027_conference}
\bibliographystyle{iclr2027_conference}

\appendix
\section{Appendix}
\label{sec:Appendix}
\subsection{Project Website}
We provide link to our project website at \url{https://fracgen.github.io/}.

\subsection{Pre-processing dynamic physics map}
FracSim \ref{subsec:FracSim} produces a set
of per-frame 2D physics maps: three scalar fields (stress, strain, damage) and one vector field (flow). Before VAE encoding, each raw map is normalized per video and linearly rescale (clipping) into $[0,1]$. For the scalar fields (stress, strain, damage), the normalized scalar is replicated identically across the three RGB channels, yielding a grayscale RGB video in which pixel intensity encodes physical magnitude; background pixels normalize to exactly zero and render as black. For the flow field, which is a 2D vector rather than a scalar, we instead apply a Middlebury-style optical-flow color-wheel encoding~\citep{4408903}. Zero flow therefore also renders as black, consistent with the convention used for the scalar fields (Note that this is what the model sees during training, for visualization as shown in Fig.\ref{fig:phys_map_pred}, we map it back to white background for the ease of visualization). Since the 3D VAE expects RGB input, each physics map is now a standard RGB video after this pre-processing, requiring no architectural change to accommodate a new modality. We feed each pre-processed physics-map video into the frozen 3D VAE to obtain its latent, as described in \S\ref{subsec:frac_gen}.
\subsection{Additional Background in Material Point Methods}
Material Point Method (MPM) \citep{jiang2016material} is a hybrid
Eulerian--Lagrangian discretization for continuum mechanics. MPM represent a body with a set of Lagrangian material points (particles), carrying position $\mathbf{x}_p$, velocity $\mathbf{v}_p$, mass $m_p$, and deformation gradient $\mathbf{F}_p$. The deformation gradient $\mathbf{F}_p$ tracks the particle's accumulated deformation, encoding rotation and strain which mainly drives the constitutive strain-stress response. At each MPM simulation step, mass and momentum are transferred to grid nodes (P2G); grid velocities are updated under internal stress and
external constraints; the updated velocities are interpolated back to the particles for advection (G2P); and each particle’s deformation gradient is evolved accordingly
\paragraph{Particle-to-grid (P2G) transfer.} Mass and momentum are accumulated onto the grid with
the affine particle-in-cell (APIC) scheme \citep{jiang2015affine}, augmenting each particle's velocity with a
local affine velocity term $\mathbf{C}_p^t$:
\begin{align}
m_i^t &= \sum_p w_{ip}^t\, m_p, \\
(m\mathbf{v})_i^t &= \sum_p w_{ip}^t\, m_p
\left[ \mathbf{v}_p^t + \mathbf{C}_p^t \left(\mathbf{x}_i^t - \mathbf{x}_p^t\right) \right],
\end{align}
where $w_{ip}^t$ is the B-spline weight coupling particle $p$ to grid node $i$.

\paragraph{Grid update.} Each grid node velocity is integrated forward under the net nodal force $\mathbf{f}_i$, combining internal and external forces: 
\begin{equation} 
    \mathbf{v}_i^{t+1} = \mathbf{v}_i^t + \frac{\Delta t}{m_i^t}\, \mathbf{f}_i\!\left(\mathbf{x}_i^t; \boldsymbol\theta_p\right). 
\end{equation} The force follows from a hyperelastic energy $\Psi(\mathbf{F})$, and $\theta_p$ gathers the governing material properties, Young's modulus $E$, and Poisson's ratio $\nu$.
\paragraph{Grid-to-particle (G2P) transfer.} The updated nodal velocities are interpolated back to
the particles, whose positions are then advanced, the local affine velocity term is also updated correspondingly:
\begin{equation}
\mathbf{v}_p^{t+1} = \sum_i w_{ip}^t\, \mathbf{v}_i^{t+1}, \qquad
\mathbf{x}_p^{t+1} = \mathbf{x}_p^t + \Delta t\, \mathbf{v}_p^{t+1}, \qquad
\mathbf{C}_p^{t+1} = \frac{4}{(\Delta x)^2} \sum_{i} w_{ip}^t \mathbf{v}_i^{t+1} (\mathbf{x}_i - \mathbf{x}_p^t)^T
\end{equation}
\paragraph{Deformation gradient update.} Each particle's deformation gradient is updated from the
velocity gradient sampled off the grid:
\begin{equation}
\mathbf{F}_p^{t+1} = \left[ \mathbf{I}
+ \Delta t \sum_i \mathbf{v}_i^{t+1} \left(\nabla w_{ip}^t\right)^{\!\top} \right] \mathbf{F}_p^t.
\end{equation}

\subsection{Additional Simulation and Training Details}
\label{subsecsion:sim_train_details}
\paragraph{From Cauchy stress tensor to von Mises scalar.} We first compute the deviatoric Cauchy stress tensor $\bm{s}_p = \tilde{\boldsymbol{\sigma}}_p - \frac{1}{3}\operatorname{tr}(\tilde{\boldsymbol{\sigma}}_p)\mathbf{I}$ and collapse it into a scalar von Mises equivalent stress $\tilde{\boldsymbol{\sigma}}_{\mathrm{vM},p} = \sqrt{\frac{3}{2} \bm{s}_p : \bm{s}_p}$.

\paragraph{Relation to elastoplasticity.} Our two material classes differ in their hyperelastic energy density (fixed corotated vs.\ neo-Hookean), elastic constants, and damage thresholds: FracSim couples hyperelasticity to continuum damage, with no yield surface and no return mapping ($F^{P}=I$). This is deliberate, as damage thresholds expose fracture timing as a directly controllable quantity, which is what FracGen conditions on. The cost is that we reproduce the visual distinction between materials that tear at low stretch and those that sustain large stretch, but not the permanent deformation of true ductile (plastic) fracture.

\paragraph{Simulation Details.} Following PhysGaussian \citep{Xie_2024_CVPR}, we optimize 3DGS per object using an anisotropy regularizer and internal particle filling. Fracture is produced by a continuum damage model coupled to MPM
(\S\ref{subsec:FracSim}), and diversity is obtained by sampling loading, constitutive model, material properties constants, and fracture properties. Every simulation is fully specified by a single configuration, which makes each sample reproducible. All runs use $100$ frames with $100$ MPM substeps per frame ($\Delta t=10^{-4}$\,s), a $150^3$ background grid over the normalized domain $[0,2]^3$, and no gravity, so that the observed deformation is attributable solely to the prescribed load. Across combinations of object, loading, and material/fracture parameters, FracSim yields a paired dataset of RGB fracture videos and dynamic physics maps for generative training.
\begin{enumerate}[label=\roman*)]
\item \textbf{Uniaxial Opposing Forces:} Tensile loading is applied kinematically. We select two thin slabs of MPM particles on opposite sides of the object center (thickness $0.05$ in simulation units, i.e.\ roughly one grid cell, and wide enough to span the object's cross-section) and prescribe their velocities to $\pm v\,\hat{\mathbf{u}}$ for the entire clip, where $\hat{\mathbf{u}}$ is the tearing axis. Because the velocity is enforced on particles rather than on grid nodes, the grips move rigidly with the material and the remaining particles respond through the constitutive law. We vary the speed $v\in[0.2,1.5]$, the tearing axis over $12$ directions in the image plane ($0^\circ$ to $165^\circ$ in $15^\circ$ steps), the grip offset from the center ($0.10$--$0.25$), and the uniaxial tensile stretch mode.
\item \textbf{Constitutive Model:} We match the constitutive model to the deformation regime each material reaches before it fails. For low-stretch objects like bread, we use the fixed-corotated model, which is accurate and stable at small-to-moderate strain. For high-stretched object like rubber toys, it sustain stretches of $3\times$ or more before failing, which requires a genuinely finite-strain energy density; for these we use the compressible neo-Hookean model, a standard model of rubber elasticity.
\item \textbf{Material Properties:} Young's modulus $E$ and Poisson's ratio $\nu$ are sampled per class (Tab. \ref{tab:sim_params}). Larger $E$ gives sharper, more localized cracks with faster snap-back; smaller $E$ gives broader necking before the tear; $\nu$ controls how much the cross-section thins under tension. Rubber objects uses much weaker velocity damping than the bread-like objects so its stored elastic energy is released visibly on fracture.
\item \textbf{Fracture Properties:} We augment each particle with a scalar damage $d_p\in[0,1]$ ($0$ intact, $1$ broken) \citep{Lematre1985ACD}. After each substep we take the principal stretches of $\mathbf{F}_p$ (its singular values) and update damage from the largest one as discussed in Eq.\ref{eq:damage} and degrade the particle's stress contribution to the grid. The onset stretch $\lambda_{\text{onset}}$ sets how far material can stretch before weakening; while the crit stretch indicates the critical stretch. Varying these thresholds independently of $E$ and $\nu$ decouples \emph{when} an object breaks from \emph{how} it deforms beforehand, and is the main source of qualitative diversity in fracture patterns.
\end{enumerate}

\begin{table}[h]
\centering\small
\caption{Per-class simulation parameters.}
\label{tab:sim_params}
\begin{tabular}{lcc}
\toprule
 & \textbf{Low-stretch objects} & \textbf{High-stretch objects} \\
\midrule
\textbf{Constitutive model} & Fixed corotated & Neo-Hookean \\
\textbf{Young modulus} $E$ & $\{2\times10^3$, $6\times10^3$, $10^4\}$ & $\{2\times10^4$, $6\times10^4$, $10^5\}$ \\
\textbf{Poisson ratio} $\nu$ & $\{0.2, 0.25, 0.3\}$ & $\{0.45, 0.47, 0.49\}$ \\
\textbf{Onset, critical stretch} $(\lambda_{\text{on}}$, $\lambda_{\text{cr}}$) & $\{(1.5,2.3), (2.0,3.0)\}$ & $\{(2.6, 3.0), (3.0, 3.3)\}$ \\
\textbf{RPIC damping / grid velocity decay} & $0.5$ / $0.9995$ & $0.005$ / $0.9998$ \\
\textbf{Grip speed} $v$ & $\{0.5, 0.6, 0.7,0.8\}$ & $\{0.6, 0.8, 1.2,1.4\}$ \\
\midrule
\textbf{Tearing axes} & \multicolumn{2}{c}{$\{0^\circ,15^\circ,\dots,165^\circ\}$} \\
\textbf{Grip offset from center} & \multicolumn{2}{c}{$\{0.10,0.15,0.20,0.25\}$} \\
\textbf{Density / gravity} & \multicolumn{2}{c}{$200$ / $0$} \\
\bottomrule
\end{tabular}
\end{table}

\begin{figure}
    \centering
    \includegraphics[width=\linewidth, trim={0cm 3cm 0cm 5cm},clip]{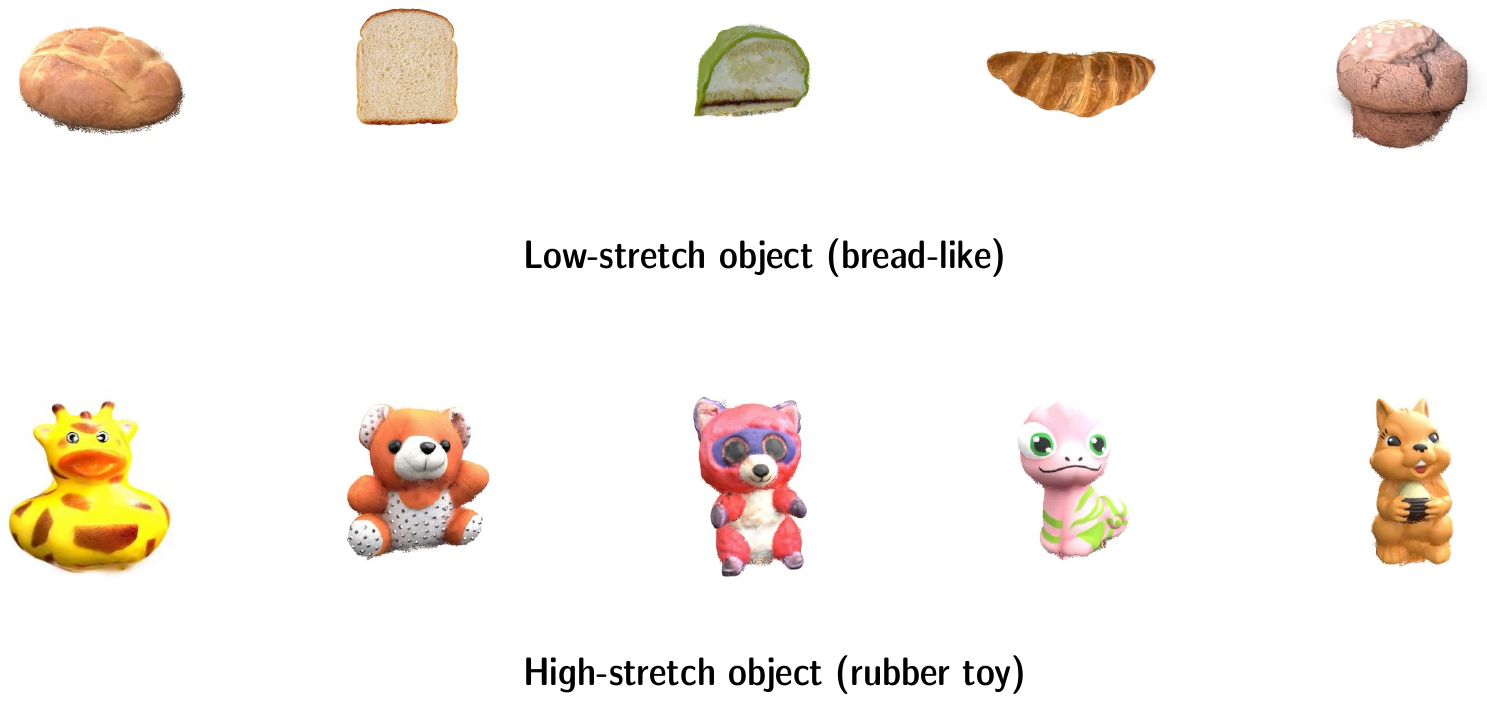}
    \caption{Ten objects used for simulation. First 3 of each category is then used for training while last two are used for testing and benchmarking.}
    \label{fig:app_10_objects}
\end{figure}

\begin{figure}
    \centering
    \includegraphics[width=\linewidth, trim={0cm 3cm 0cm 5cm},clip]{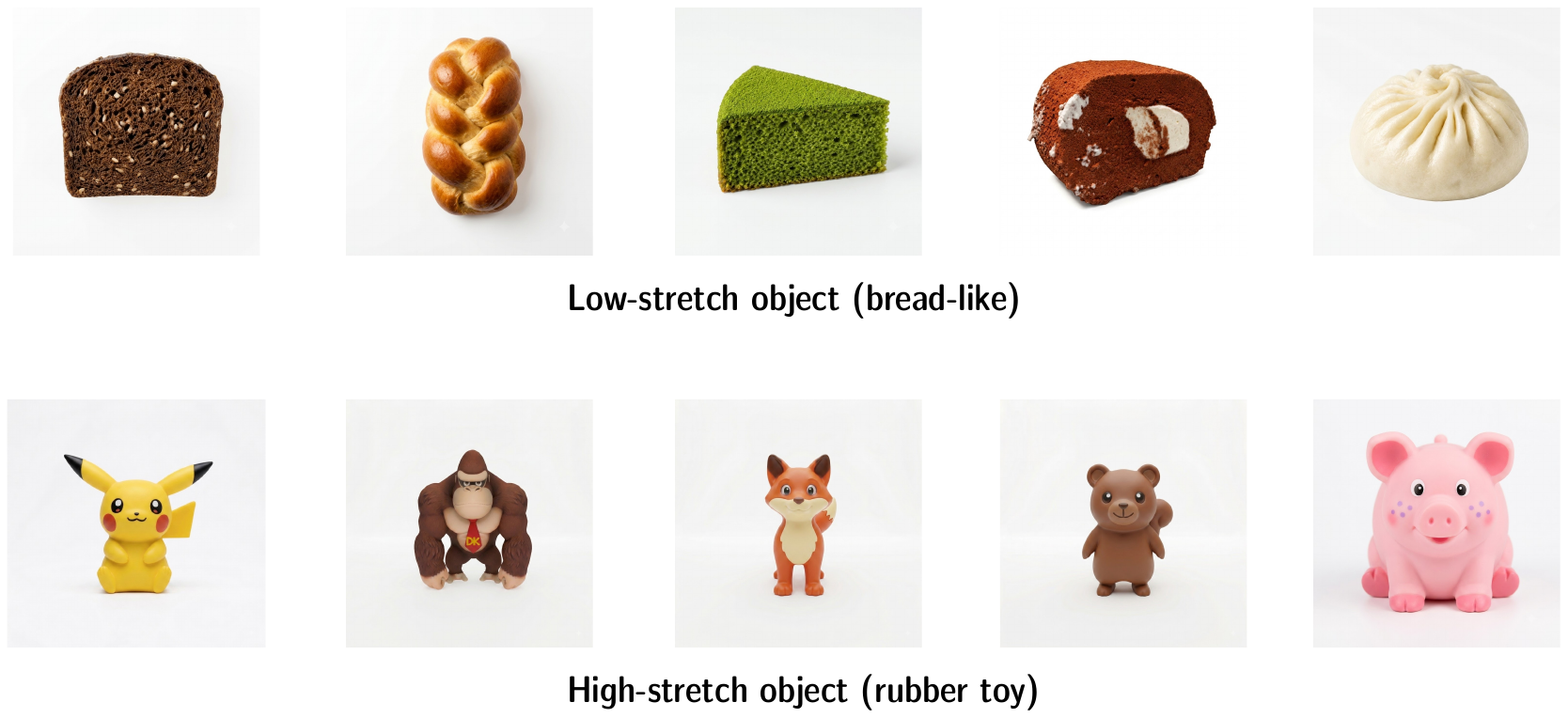}
    \caption{More ten random objects used for testing FracGen generalization. Results can be found on our website.}
    \label{fig:app_10_random_objects}
\end{figure}

\paragraph{Training Details.}
We leverage simulated data from FracSim to train our FracGen model. Among 10 objects, simulated results from 6 objects are used for training (3 low-stretch and 3 high-stretch) and 4 objects are used for testing, and benchmarking (2 low-stretch and 2 high-stretch). Our model is finetuned from a pre-trained Wan 2.1-Fun-V1.1-1.3B-Control \citep{wan2025wan} using LoRA \citep{hu2022lora} with rank 64, applied to the query, key, value, output, and feed-forward projections of the DiT backbone. We use the AdamW optimizer with a learning rate of 1e-4, weight decay of 0.01, and a cosine learning-rate schedule with a 40-step warmup. Each video clip consists of 81 frames, and the model is trained with a per-GPU batch size of 1 and gradient accumulation of 4 steps across 4 GPUs, yielding an effective batch size of 16. The total training samples are 4,212 training samples, and total testing/benchmarking samples are 2,808. Additionally, we also collect 10 random object images to further test generalization of FracGen across different new objects. We show images of these objects in Fig.\ref{fig:app_10_random_objects}.

\subsection{Additional Details on Evaluation Metric Design}
\label{subsection:eval_metric_design}
\paragraph{Stretching Axis Coherence (SAC).} 

\begin{figure}
    \centering
    \includegraphics[width=1.02\linewidth, trim={0cm 5cm 0cm 3cm},clip]{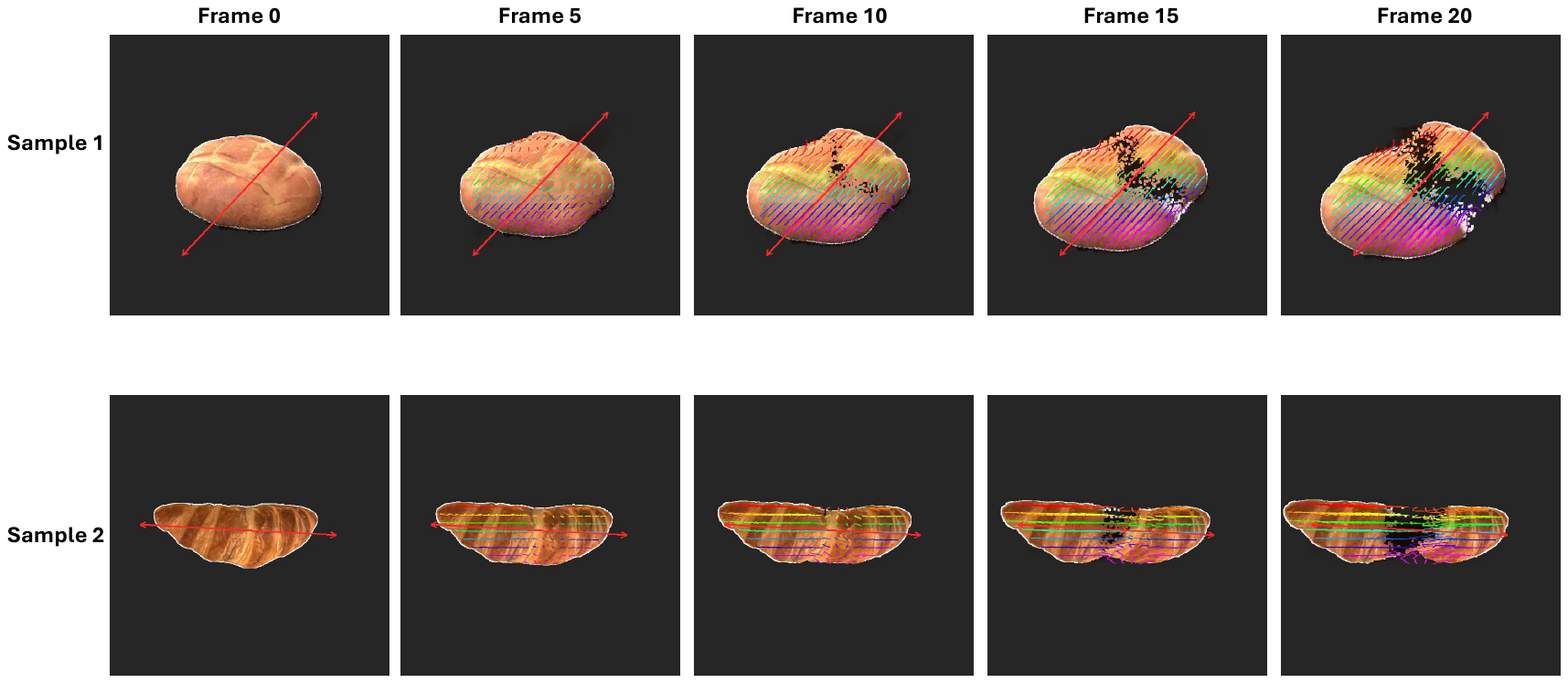}
    \caption{We show illustration of using point tracks to extract principal stretching axis.}
    \label{fig:sac_visualization}
\end{figure}

Given a fracture video, we first segment out main object using SAM \citep{kirillov2023segany} and run point track algorithm \citep{lai2026a} to extract all track information. Point track results are visualized in Fig.\ref{fig:sac_visualization}. Given 2D point tracks $\{\mathbf{p}_i(t)\}_{i=1}^{N}$, we estimate the dominant deformation axis over an
early window $[t_0,t_1]$ and compare it against the same quantity measured on a reference (ground-truth) clip.

\textit{\textbf{Displacement and drift removal.}}
For each point $i$ visible at both $t_0,t_1$, we extract its displacement $\mathbf{u}_i=\mathbf{p}_i(t_1)-\mathbf{p}_i(t_0)$. The mean
$\bar{\mathbf{u}}=\tfrac{1}{N}\sum_i\mathbf{u}_i$ is dominated by the
object's rigid drift, since opposing deformation components cancel in the
average. We subtract this quantity from each point's displacement, giving deformation-relative motion
$\hat{\mathbf{u}}_i=\mathbf{u}_i-\bar{\mathbf{u}}$.

\textit{\textbf{Noise gating}.}
Point tracks might be unstable due to noisy prediction from tracking algorithm or other noisy motion. We use point's displacement magnitude $w_i=\lVert\hat{\mathbf{u}}_i\rVert$ as a gating function to drop these noisy points with $w_i < \tau$. If fewer than $k$ points survive, we determine the axis to be undefined (the object has not yet shown measurable deformation).

\textit{\textbf{Dominant axis extraction}.}
Since opposite sides of a stretching object move in opposite directions, averaging directions directly would cancel to zero. We instead form the orientation tensor over unit directions $\mathbf{e}_i=\hat{\mathbf{u}}_i/w_i$:
\begin{equation}
    Q = \frac{\sum_i w_i\, \mathbf{e}_i \mathbf{e}_i^{\top}}{\sum_i w_i}
    \in \mathbb{R}^{2\times 2},
    \label{eq:sac-Q}
\end{equation}

Intuitively, the outer product $\mathbf{e}_i \mathbf{e}_i^{\top}$ does not care about direction and hence $\mathbf{e}\mathbf{e}^{\top} = (-\mathbf{e})(-\mathbf{e})^{\top}$. Weighted average of these terms give the overall voting principal axis. We then perform eigen-decompose on the symmetric matrix $Q$ to extract orthogonal eigenvectors with eigenvalues $\lambda_{\max} \ge \lambda_{\min}
\ge 0$; the eigenvector of $\lambda_{\max}$ is the dominant axis $\mathbf{a}$. We show in Fig.\ref{fig:sac_visualization} the principal stretching axis (overlaying red line) computed using this approach from two samples of our fracture video. We also determine a confidence score,
\begin{equation}
    c_{\text{pred}} = 1 - \frac{\lambda_{\min}}{\lambda_{\max}} \in [0,1]
    \label{eq:sac-confidence}
\end{equation}
Intuitively, $c_{\text{pred}} \to 1$ when motion is tightly aligned along one line, $c_{\text{pred}} \to 0$ when motion is isotropically scattered.

\textit{\textbf{Computing SAC.}}
Given the principal stretching axis of prediction and groundtruth fracture video $\mathbf{a}_{\text{pred}}, \mathbf{a}_{\text{gt}}$, we measure their angular
deviation via
\begin{equation}
    \theta = \arccos\!\big(\,|\mathbf{a}_{\text{pred}} \cdot \mathbf{a}_{\text{gt}}|\,\big)
    \in [0^{\circ}, 90^{\circ}],
    \label{eq:sac-angle}
\end{equation}
where the absolute value dropping the axis sign ambiguity. This is squashed
into a bounded similarity score via a Gaussian kernel,
\begin{equation}
    s_{\text{raw}} = \exp\!\left(-\left(\theta / \sigma\right)^2\right) \in (0, 1],
    \label{eq:sac-rawscore}
\end{equation}
In this way, a perfect match ($\theta = 0$) scores $1$ and the score decays smoothly
with angular error (we set $\sigma = 15^{\circ}$). The final SAC score
additionally is computed as,
\begin{equation}
    \mathrm{SAC} = s_{\text{raw}} \cdot c_{\text{pred}},
    \label{eq:sac-final}
\end{equation}
with $c_{\text{pred}} := 0$ whenever the predicted axis itself is undefined
(scored as a hard failure, $\mathrm{SAC} = 0$, rather than excluded). 
\paragraph{Damage Progression Alignment (DPA).} Let $d(\mathbf{x},t)\in[0,1]$ denote the damage field of a clip with $T$ frames and $H\times W$ pixels, read from its damage-map video. For each frame we record its total damage, normalized by frame size,
\begin{equation}
    a(t)=\frac{1}{HW}\sum_{\mathbf{x}}d(\mathbf{x},t),
    \label{eq:dpa-curve}
\end{equation}
together with the peak damage the clip ever attains,
$a^{\star}=\max_{t}a(t)$. The idea of DPA is to assess whether the prediction and the ground truth reach the same damage checkpoints at the same moments. A checkpoint is a fraction $\beta$ of a clip's own peak damage, and the moment at which the clip reaches it is
\begin{equation}
    f_\beta=\min\big\{t:\ a(t)\ \ge\ \beta\,a^{\star}\big\},
    \qquad
    \tau_\beta=\frac{f_\beta}{T}.
    \label{eq:dpa-crossing}
\end{equation}
With $\Delta\tau_\beta=\tau^{\text{pred}}_\beta - \tau^{\text{gt}}_\beta$, DPA is the normalized agreement averaged over a set of checkpoints $B$:
\begin{equation}
    \mathrm{DPA}=\frac{1}{|B|}\sum_{\beta\in B}
    \exp\!\Big(-\big(\Delta\tau_\beta/\sigma\big)^{2}\Big)\in(0,1],
    \qquad \sigma=0.1 .
    \label{eq:dpa-score}
\end{equation}
We sweep through a set of $B=\{0.1,0.2,0.3,0.4,0.5,0.6,0.7,0.8,0.9\}$.
\paragraph{Constitutive Consistency ($R^2_{\text{cons}}$).}
Let $\sigma(\mathbf{x},t)$, $\varepsilon(\mathbf{x},t)$ and $d(\mathbf{x},t)$ denote a predicted stress, strain and damage fields, read from its physics-map videos. In intact material linear elasticity requires stress to be proportional to strain; once damage accumulates the material softens, so stress falls while strain continues to grow. Therefore, we restrict attention to material that has not yet failed,

\begin{equation}
    S=\big\{(\mathbf{x},t):\ d(\mathbf{x},t)\le d_{\max}\big\},
    \qquad d_{\max}=0.05,
    \label{eq:cons-mask}
\end{equation}
On $S$ we fit a single slope through the origin --- zero strain must produce zero stress --- and record the coefficient of determination,
\begin{equation}
    k=\frac{\sum_{S}\varepsilon\,\sigma}{\sum_{S}\varepsilon^{2}},
    \qquad
    R^2_{\text{cons}}=1-\frac{\sum_{S}\big(\sigma-k\,\varepsilon\big)^{2}}
                             {\sum_{S}\sigma^{2}} .
    \label{eq:cons-score}
\end{equation}
$R^2_{\text{cons}}$ therefore measures how tightly a single linear stress--strain law explains the maps. To validate the robustness of this metric, we evaluate the identical quantity on the ground-truth maps, obtaining $\mathbf{R}^2_{\text{cons}}=0.948$, which serves as the upper bound. Against it, our FracGen (with constitutive loss) attains $0.902$ (95\% of the ceiling) while the variant without constitutive loss reaches only $0.378$, indicating that the latter's predicted stress and strain fields are not governed by any single elastic law.
\subsection{Additional Details on Building Annotation Tool}
\begin{figure}
    \centering
    \includegraphics[width=1.02\linewidth, trim={0cm 5cm 0cm 3cm},clip]{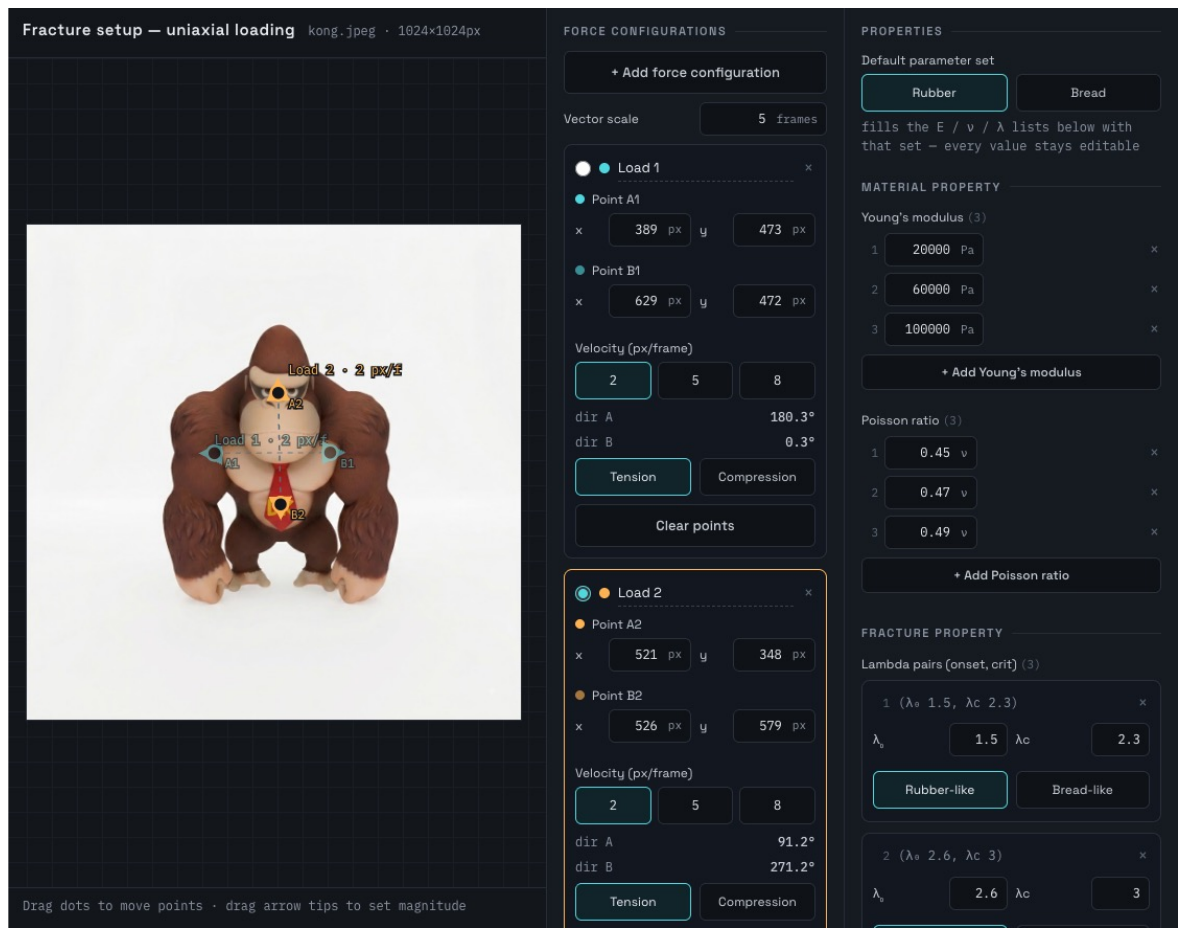}
    \caption{Screenshot of annotation tool to provide physics condition for FracGen}
    \label{fig:annotation_tools}
\end{figure}

For comprehensive annotating force applied on the single image object, we build an annotation tool that takes an input image and allow user to annotate physics condition. Specifically, users are given a UI to specify uniaxial applied load, material properties, and fracture properties. We show a screenshot of this annotation tool in Fig.\ref{fig:annotation_tools}.

\subsection{Additional Details on Baseline Implementation}
\label{subsec:baseline_details}
We provide more details on our attempt on fine-tuning baseline for fair comparison as below.
\paragraph{Matched training protocol.} All fine-tuned baselines are trained on the identical FracSim corpus and train/test split used for FracGen (\S\ref{sec:experiments}). Rather than using FracGen's hyperparameters on backbones of differing scale, each model is trained under its default setting described in its original work. When a baseline exposes a conditioning channel for applied force we use it, adapted to our two-grip setup as described below; when it does not, force is described in the text prompt only, which cannot express pull direction or magnitude.

\begin{enumerate}
    \item \textbf{ForcePrompting} \citep{gillman2026force} fine-tunes a ControlNet on CogVideoX-5B-I2V to follow a rendered Gaussian-blob control video encoding a single applied force (a point poke or a global wind direction). Since tearing requires two opposing pulls rather than one, we extend its point-force conditioning with a custom \emph{double-force} variant that renders and sums two independent Gaussian blobs at the two grip locations, matching the force representation FracGen consumes. We then finetune this ControlNet on our FracSim force-blob/RGB pairs under the matched protocol. The backbone remains frozen, following the original method. There are no conditions on other channels such as Young's modulus, Poisson's ratio, or fracture thresholds, so this baseline sees the same grips and the same pull velocities as FracGen but cannot be told what the object is made of.
    \item \textbf{PhyCo} \citep{narayanan2026phyco} conditions a frozen Cosmos-Predict2 Video2World backbone with a ControlNet over friction, restitution, deformability, and a single static applied force point. We adapt its data pipeline to accept two independent force points rather than one, to represent our two-grip pulling setup. We then finetune their ControlNet on FracSim data under the same matched training control.
    \item \textbf{CogVideoX-5B-I2V} \citep{yang2025cogvideox} has no mechanism for force or physics conditioning at all. Hence, we use intact-object as first frame condition and use a caption to describe tearing behavior (e.g.\ ``Two opposing forces pull on the object, tearing it apart at the center'').
    We fine-tune it on our the same data provided by FracSim with LoRA of rank 64. Pull direction and magnitude cannot be expressed beyond this, so identical captions are used for opposing pull directions of the same object.
\end{enumerate}

After fine-tuning, every baseline was trained on the same data distribution as similar to FracGen, making it a fair comparison. We also report results for these baselines before and after fine-tuning on the data set produced by our FracSim in Tab.\ref{tab:baselines}. This shows that even with training on the same dataset, our FracGen can still outperform these fine-tuned baselines.

\subsection{Additional Results}
\begin{figure}
    \centering
    \includegraphics[width=1.02\linewidth, trim={0cm 3cm 0cm 2cm},clip]{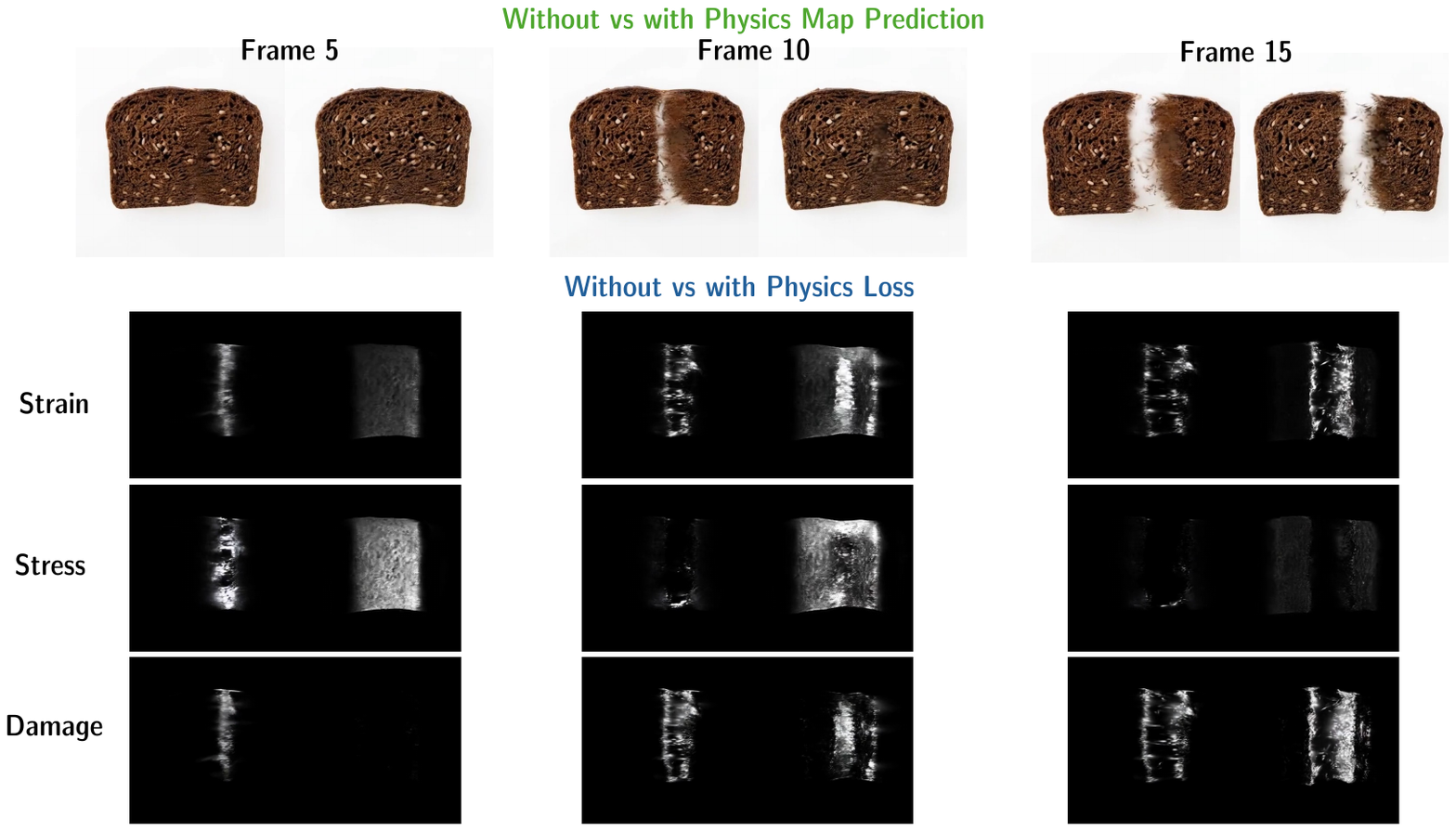}
    \caption{\textbf{Ablation Studies.} Comparison results for ablating the effect of physics map prediction and physics consistency loss.}
    \label{fig:ablate_fig}
\end{figure}

We present more comparison and controllable results on our website. Additionally, we show qualitative example for ablation studies on the effect of physics map prediction as well as physics consistency loss in Fig.\ref{fig:ablate_fig}. As shown, physics map prediction helps to improve physics fidelity of tearing bread (softening first before breaking). On the other hand, physics consistency loss helps to ensure consistency and accurate physics map prediction for stress, strain, and damage. 

\begin{table}[t]
\centering
\caption{Comparison of baselines with their zero-shot and fine-tuned (-FT) version.}
\begin{tabular}{lcccc}
\toprule
\textbf{Method} & \textbf{FVD} $\downarrow$ & \textbf{LPIPS} $\downarrow$ & \textbf{PSNR} $\uparrow$ & \textbf{SAC} $\uparrow$ \\
\midrule
ForcePrompting & 1226.32 & 0.32 & 15.07 & 0.075 \\
ForcePrompting-FT & \textbf{933.21} & \textbf{0.29} & \textbf{15.93} & \textbf{0.092} \\
\midrule
PhyCo & 1595.01 & 0.38 & 12.51 & 0.415 \\
PhyCo-FT & \textbf{832.12} & \textbf{0.33} & \textbf{14.18} & \textbf{0.434} \\
\midrule
CogVideoX-5B-I2V & 1844.92 & 0.34 & 13.78 & \textbf{0.470} \\
CogVideoX-5B-I2V-FT & \textbf{1533.21} & \textbf{0.32} & \textbf{14.12} & 0.432 \\
\bottomrule
\end{tabular}
\label{tab:baselines}
\end{table}
\subsection{More Related works}
\paragraph{3D Gaussians Splatting.} 3D Gaussian Splatting (3DGS) \citep{kerbl3Dgaussians} represents a static 3D scene as a set of anisotropic Gaussian kernels $\{G_p\}_{p=1}^{\mathcal{P}}$. Each Gaussian $G_p = \{x_p, \sigma_p, A_p, \mathcal{C}_p\}$ carries a center $x_p$, opacity $\sigma_p$, covariance $A_p$, and spherical harmonic coefficients $\mathcal{C}_p$. For rendering, the Gaussians are projected onto the image plane and composited front-to-back:
\begin{equation}
G_p = \{x_p, \sigma_p, A_p, C_p\}, \qquad
C = \sum_{p \in \mathcal{P}} \alpha_p \, \mathrm{SH}(l_p; \mathcal{C}_p) \prod_{j=1}^{p-1} (1 - \alpha_j).
\label{eq:rendering}
\end{equation}

where $\alpha_p$ is the effective opacity at pixel location, $\mathrm{SH}(l_p; \mathcal{C}_p)$ is the view-dependent color given view $l_p$, and the product term is the accumulated transmittance. The Gaussians are optimized via photometric loss across multiple views. We adopt the original 3DGS framework to reconstruct static objects, which are then passed to FracSim (\S\ref{subsec:FracSim}) to simulate fracture dynamics.
\paragraph{Ductile Fracture Mechanics.}
Dynamic fracture~\citep{Kachanov1999,Lematre1985ACD,wolper2019cd,Patnaik2021} is among the most challenging phenomena to model in continuum mechanics. A central concept is material \emph{ductility}---the capacity to deform plastically before failure---which separates the two canonical fracture modes. \emph{Ductile} fracture occurs in materials that sustain large deformation under load, typically necking before the cross-section separates, whereas \emph{brittle} fracture occurs at very low strains, often below $5\%$, with little observable deformation prior to failure. In this work, we target the kinematic regime of ductile fracture (large deformation before failure), modeled via hyperelasticity with stretch-driven damage rather than plasticity. Specifically, we study fracture under uniaxial tensile loading, where two opposing forces act along a common axis, causing the object to stretch, deform, and eventually break. Under load, the body extends along the loading (longitudinal) direction while contracting in the transverse directions. Prior work on fracture simulation achieves high physical fidelity but requires expert knowledge to configure the governing parameters. CD-MPM \citep{wolper2019cd} models dynamic ductile fracture through continuum damage mechanics, coupling a phase-field damage evolution with the MPM to capture large-deformation crack propagation. To reduce the heavy runtime and memory cost of full-order solvers, Neural Stress Fields (NSF) \citep{zong2023neural} learns a low-dimensional manifold of the Kirchhoff stress field through an implicit neural representation, enabling reduced-order simulation of elastoplasticity and fracture. These methods, however, target physical simulation or geometric fragmentation, and expose control only through low-level physical parameters. To the best of our knowledge, our work is the first to bring stretch-to-tear fracture to a generative model---teaching a video generator the full deformation-to-fracture progression under comprehensive input conditioning, so that users can control which object to break, when it should break, how it breaks. FracSim captures the kinematic signature of this regime — large stretch, necking, progressive separation — through damage accumulation rather than plastic flow.

\end{document}